\documentclass[10pt,twocolumn,letterpaper]{article}

\usepackage[pagenumbers]{cvpr}      

\usepackage[T1]{fontenc}
\usepackage[stretch=30,shrink=30]{microtype}
\newcommand{\method}{\textsc{QuantAtlas-OM}\xspace}

\definecolor{geometrygreen}{HTML}{397E76}
\definecolor{geometryorange}{HTML}{B87747}

\usepackage{colortbl}
\usepackage{tabularx}
\usepackage{pifont}
\definecolor{comparisonlavender}{HTML}{F2EEF9}
\definecolor{comparisonpositive}{HTML}{16A05D}
\definecolor{comparisonnegative}{HTML}{D43838}
\definecolor{comparisonpartial}{HTML}{9A5B08}

\usepackage{tikz}

\usepackage{cuted}
\newcommand{\OceanTwoColumn}{}

\definecolor{cvprblue}{rgb}{0.21,0.49,0.74}
\usepackage[pagebackref,breaklinks,colorlinks,allcolors=cvprblue]{hyperref}

\def\confName{CVPR}
\def\confYear{2027}

\title{QuantAtlas-OM: Taming Neural Spectral Bias in Long-Horizon Spatiotemporal Ocean Dynamics}
\author{\href{https://openreview.net/profile?id=\string~Hao_Wu39}{\textcolor{black}{Hao Wu}}\textsuperscript{1,7,$*,\dagger$}\quad \href{https://openreview.net/profile?id=\string~Yuan_Gao26}{\textcolor{black}{Yuan Gao}}\textsuperscript{1,7,$*,\dagger$}\quad Tong Wu\textsuperscript{2}\quad \href{https://openreview.net/profile?id=\string~Fan_Xu5}{\textcolor{black}{Fan Xu}}\textsuperscript{3}\quad \href{https://openreview.net/profile?id=\string~Fan_Zhang27}{\textcolor{black}{Fan Zhang}}\textsuperscript{4}\quad \href{https://openreview.net/profile?id=\string~Jinhan_Li4}{\textcolor{black}{Jinhan Li}}\textsuperscript{3}\\[2pt]
\href{https://openreview.net/profile?id=\string~Kun_Wang15}{\textcolor{black}{Kun Wang}}\textsuperscript{5}\quad \href{https://openreview.net/profile?id=\string~Zhenyu_Ma5}{\textcolor{black}{Zhenyu Ma}}\textsuperscript{2}\quad \href{https://openreview.net/profile?id=\string~Bryan_Dai1}{\textcolor{black}{Bryan Dai}}\textsuperscript{2}\quad \href{https://openreview.net/profile?id=\string~Xiaomeng_Huang1}{\textcolor{black}{Xiaomeng Huang}}\textsuperscript{1,$\dagger$}\quad \href{https://openreview.net/profile?id=\string~Qingsong_Wen2}{\textcolor{black}{Qingsong Wen}}\textsuperscript{6,$\dagger$}\\[5pt]
{\small \textsuperscript{1}THU\quad \textsuperscript{2}IQuest Research\quad \textsuperscript{3}USTC\quad \textsuperscript{4}CUHK\quad \textsuperscript{5}NTU\quad \textsuperscript{6}Squirrel Ai Learning\quad \textsuperscript{7}Ubiquant Investment}\\[2pt]{\small \textsuperscript{*}Equal contribution.\qquad\textsuperscript{$\dagger$}Corresponding authors.}}
\hypersetup{pdftitle={QuantAtlas-OM: Taming Neural Spectral Bias in Long-Horizon Spatiotemporal Ocean Dynamics},pdfauthor={Hao Wu, Yuan Gao, Yu Xu, Fan Xu, Fan Zhang, Jinhan Li, Kun Wang, Zhenyu Ma, Bryan Dai, Xiaomeng Huang, Qingsong Wen}}


\begin{document}
\maketitle
\begin{abstract}
Autoregressive neural networks exhibit spectral bias in long-horizon spatiotemporal prediction. As rollouts progress, high-frequency details decay, blurring the predicted fields and destabilizing long-range forecasts.
Ocean dynamics offer a particularly stringent test because kinetic energy is concentrated in mesoscale eddies and fronts spanning tens of kilometers.
This degradation stems from coupling large-scale phase evolution and small-scale energy retention in a shared feature space. Evolution on a coarse grid smooths out high-wavenumber eddies, whereas evolution on a fine grid amplifies small-scale phase errors and triggers rollout instability.
We introduce \method (\textbf{QuantAtlas Ocean Model}), a scale-spectral decoupling framework that confines deep temporal evolution to a low-wavenumber latent manifold for long-term stability while routing full-resolution eddy and front residuals through an outer bypass to the decoder.
Across physical regimes, the evolution core pairs non-local attention with resolution-invariant dilated spatial gating.
Across three western boundary currents, \method retains $43.7\%$ to $86.5\%$ of mesoscale spectral energy along the $k^{-3}$ cascade at Day~$120$, sustains anomaly correlation coefficients of $0.81$ to $0.88$ at Day~$60$, and reaches $0.66$ at Day~$120$ in the Kuroshio ($2.5\times$ to $7.7\times$ the best baselines).
Trained at $0.25^\circ$, \method transfers zero-shot to a global $0.125^\circ$ grid ($0.81$ Day-$10$ spatial correlation) and leads global $0.25^\circ$, $23$-level 3D ocean simulations ($120$ days) and coupled ocean-atmosphere forecasts ($60$ days), ranking first on $15$ of $16$ metrics and cutting $60$- to $120$-day salinity RMSE by $13\%$ to $27\%$ across global ocean basins.
\end{abstract}

\section{Introduction}
\label{sec:intro}
\begingroup
\hyphenpenalty=10000
\exhyphenpenalty=10000
\emergencystretch=1em

\begin{figure}[t]
  \ifdefined\OceanTwoColumn\vspace*{24pt}\fi
  \centering
  \includegraphics[width=\columnwidth]{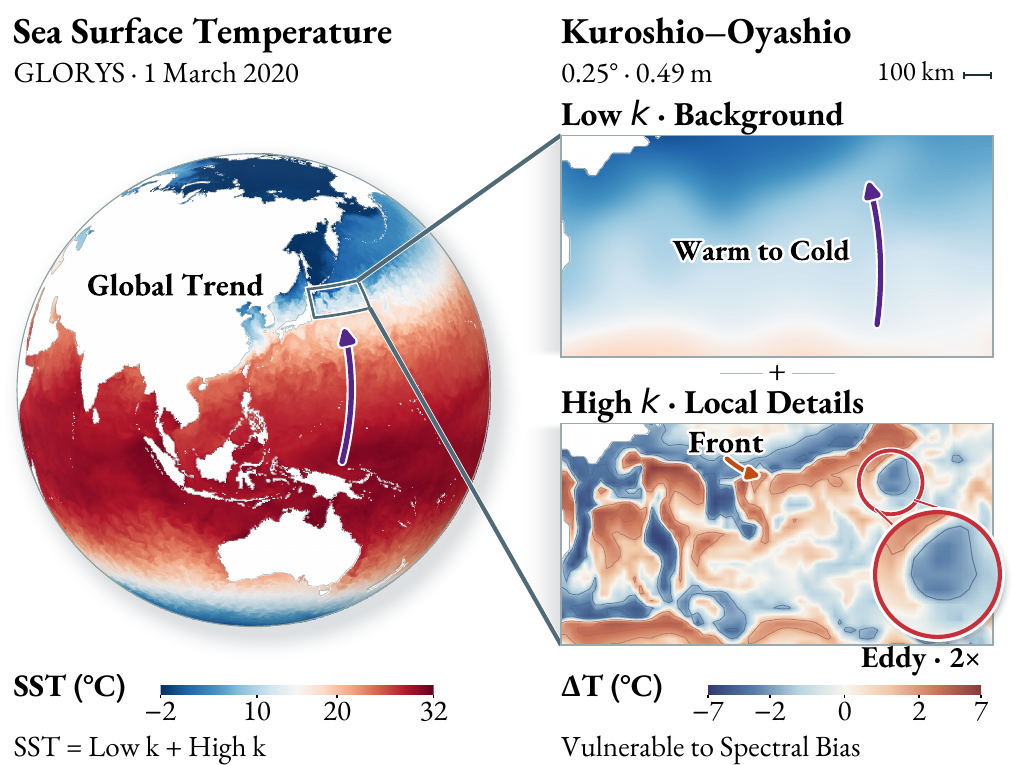}
  \caption{\textbf{\textit{Global trends and local details.}}
  GLORYS sea surface temperature~\cite{lellouche2021glorys} separates into a smooth background and high-frequency frontal and eddy residuals. These local details are vulnerable to spectral bias during long autoregressive rollouts.}
  \label{fig:ocean_intro}
\end{figure}

Autoregressive prediction over long horizons is a central problem in spatiotemporal physical modeling~\cite{sanchez2020simulation,bonev2023sfno,wang2024nuwadynamics}. Starting from the current state, a model recursively predicts future states over tens of days or months~\cite{lam2023graphcast,bi2023pangu,chattopadhyay2024oceannet,gao2025oneforecast,gao2026neuralom}. Yet autoregressive models consistently encounter a fundamental limitation as rollouts progress and the predicted fields become increasingly blurred~\cite{chattopadhyay2024stability,lippe2023refiner,wu2026dino}. Fine-scale structures disappear first, followed by frontal weakening and eventual large-scale flow distortion~\cite{chattopadhyay2024oceannet}. A primary driver of this degradation is the spectral bias of neural networks~\cite{rahaman2019spectral,chattopadhyay2024stability}, since networks preferentially fit low-frequency components and attenuate high-frequency modes. Such high-frequency losses remain subtle in single-step prediction~\cite{lippe2023refiner}. Over tens or hundreds of closed-loop steps, however, each prediction feeds into the next step~\cite{sanchez2020simulation,bi2023pangu,wu2026pnpcorrector}, and high-frequency attenuation compounds exponentially across extended rollouts.

Ocean dynamics brings this challenge into sharp focus. Unlike the atmosphere, the ocean has a Rossby deformation radius of only tens of kilometers~\cite{chelton1998rossby}, with most kinetic energy concentrated in fine-scale fronts and mesoscale eddies~\cite{chelton2011eddies,storer2022energy}. As Figure~\ref{fig:ocean_intro} shows, sea surface temperature in the Kuroshio and Oyashio transition region decomposes into a smooth low-wavenumber background in the upper panel that captures basin-scale temperature patterns, and high-wavenumber frontal and eddy residuals in the lower panel. These residuals are far from negligible perturbations. Their temperature variations reach approximately $\pm 7\,^{\circ}\mathrm{C}$, comparable to the temperature contrasts in the background field, yet their spatial extent is only about $100\,\mathrm{km}$, equivalent to just a few grid cells at $0.25^{\circ}$ resolution. These high-frequency residuals govern ocean heat transport, cross-scale mixing, and extreme events such as marine heatwaves~\cite{dong2014transport,bian2023heatwaves}. Under spectral bias, however, neural networks fit the smooth background in the upper panel while progressively erasing the high-frequency frontal and eddy residuals in the lower panel across multi-month autoregressive rollouts~\cite{chattopadhyay2024oceannet,lippe2023refiner}.

A direct attempt to preserve eddies spanning only a few grid cells is to evolve features directly on high-resolution maps. Our experiments, however, reveal a \emph{spectral dilemma} when evolution resolution alone changes. Downsampling the fields for temporal evolution on a coarse grid truncates high-frequency residuals during encoding, irreversibly blurring eddies and fronts. Conversely, long-horizon evolution directly on a fine grid amplifies small-scale, high-frequency phase errors through autoregressive steps and propagates them toward larger scales, destabilizing the flow field. On a single feature manifold, coarse-grid evolution preserves stability at the expense of high-frequency detail, whereas fine-grid evolution retains detail at the cost of long-term rollout stability and physical consistency.

The decomposition in Figure~\ref{fig:ocean_intro} points to a clear structural resolution. Because the smooth background and high-frequency residuals differ fundamentally in spatial scale and dynamical behavior, a single deep evolution module should not carry both. Based on this insight, we propose \method (\textbf{QuantAtlas Ocean Model}), a \emph{framework for scale and spectral decoupling} (Figure~\ref{fig:ocean_method}). The encoder separates the ocean state into two orthogonal pathways. A pathway through a coarse latent grid filters out high-frequency truncation noise through strided downward projection and applies deep temporal evolution exclusively to the low-wavenumber background flow, anchoring long-term phase stability. Unlike multiscale U-Net architectures that inject intermediate skip connections into deeper layers and reintroduce high-frequency perturbations into temporal evolution, we retain only a single outer bypass at full resolution. This bypass routes frontal and eddy residuals around the entire evolution bottleneck and directly to the decoder, where the evolved large-scale flow guides their local realignment and refinement. To match the computational and transfer requirements of different ocean regimes, the latent evolution core provides two spatial mixing operators. On the $1/4$ coarse grid, a $16\times$ reduction in spatial tokens makes non-local self-attention~\cite{vaswani2017attention} tractable for capturing long-range coupling across depth levels and basins in three-dimensional baroclinic oceans and energetic regional currents. For global surface eddy fields, a multiscale dilated spatial gating operator built entirely from local and dilated depthwise convolutions~\cite{yu2016dilated,guo2023van} decouples the learned kernels from fixed grid dimensions, naturally enabling zero-shot cross-resolution transfer across ocean grids.

We evaluate \method on global $0.25^{\circ}$ 3D ocean dynamics ($23$ levels, $93$ channels), $120$-day regional $0.125^{\circ}$ mesoscale eddy forecasting across the Kuroshio, Gulf Stream, and Agulhas Current, and zero-shot $0.25^{\circ}\to0.125^{\circ}$ global transfer. Decoupling high- and low-frequency evolution retains $43.7\%$ to $86.5\%$ of $100$ to $250\,\mathrm{km}$ mesoscale energy at Day~$120$ along the $k^{-3}$ cascade~\cite{charney1971turbulence} and cuts Day-$60$ eddy RMSE by $48\%$ to $58\%$.

Our contributions are threefold.
\ding{182}~\textcolor{black}{\textbf{\textit{Problem Insight.}}} We identify a spectral dilemma in long-horizon autoregressive prediction, where coarse-grid evolution dissipates high-frequency details while fine-grid evolution amplifies small-scale phase errors and destabilizes rollouts.
\ding{183}~\textcolor{black}{\textbf{\textit{Simple and Effective Methodology.}}} We propose \method, a scale-spectral decoupling framework that combines coarse-grid latent evolution with a single full-resolution bypass, pairing non-local attention and dilated spatial gating with distinct ocean dynamical regimes.
\ding{184}~\textcolor{black}{\textbf{\textit{Performance over Long Horizons.}}} \method elevates Day-$120$ regional eddy ACC by $2.5\times$ to $7.7\times$ ($0.81$ to $0.88$ at Day~$60$), reaches $0.81$ Day-$10$ spatial correlation in zero-shot $0.25^{\circ}\to0.125^{\circ}$ global transfer, and cuts $60$- to $120$-day salinity RMSE by $13\%$ to $27\%$ in global 3D ocean simulation and coupled ocean-atmosphere forecasting.
\par
\endgroup

\ifdefined\OceanTwoColumn
\begin{figure*}[t]
  \centering
  \includegraphics[width=\textwidth]{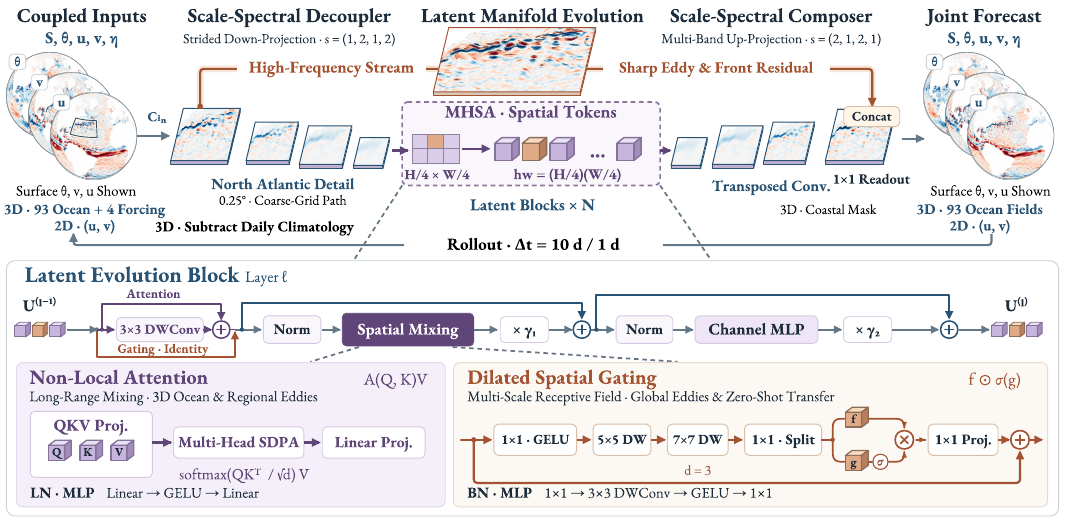}
  \caption{\textbf{\textit{\method architecture.}}
  The high-resolution stream carries fine-grid features to the decoder. On the coarse grid, the latent evolution block mixes spatial information through either non-local attention or resolution-invariant dilated spatial gating.}
  \label{fig:ocean_method}
\end{figure*}

\fi

\section{Related Work}
\label{sec:related}

\noindent\textbf{\textit{Spatiotemporal Backbones and Neural Operators.}}
Data-driven fluid modeling~\cite{wu2024prometheus,wu2024pure,wu2024pastnet} relies on spatiotemporal backbones using convolutions (SimVP~\cite{gao2022simvp}) or attention (Earthformer~\cite{gao2022earthformer}), and neural PDE operators using spectral (FNO~\cite{li2021fourier}, SFNO~\cite{bonev2023sfno}) or anti-aliased convolutions (CNO~\cite{raonic2023convolutional}). Because FNO truncates high wavenumbers and CNO routes high frequencies through intermediate skips, evolving all scales in a shared deep space smooths out fine residuals over long rollouts. \method confines deep evolution to a coarse manifold while routing fine-scale residuals through a single outer full-resolution bypass.

\noindent\textbf{\textit{Data-Driven Ocean and Spectral Bias Mitigation.}}
Global weather models (FourCastNet~\cite{pathak2022fourcastnet}, Pangu-Weather~\cite{bi2023pangu}, GraphCast~\cite{lam2023graphcast}, FuXi~\cite{chen2023fuxi}) have inspired ocean models for multivariable pretraining (AI-GOMS~\cite{xiong2023ai}), 3D circulation (ORCA-DL~\cite{guo2025orca}), eddy forecasting (WenHai~\cite{cui2025forecasting}), and regional dynamics (OceanNet~\cite{chattopadhyay2024oceannet}). Yet oceanic Rossby radii span only tens of kilometers~\cite{chelton1998rossby, storer2022energy}, and neural spectral bias~\cite{rahaman2019spectral, xu2020frequency, tancik2020fourier} couples with cross-scale cascades during long rollouts to trigger high-frequency dissipation or instability~\cite{chattopadhyay2024stability}. Unlike multi-step diffusion refinement in PDE-Refiner~\cite{lippe2023refiner} or predictor-corrector spectral regularization in OceanNet~\cite{chattopadhyay2024oceannet}, \method decouples low-wavenumber evolution from high-frequency transport in a single forward pass and supports zero-shot $0.25^\circ\to0.125^\circ$ cross-resolution transfer across global ocean grids.

\section{Method}
\label{sec:method}
\ifdefined\OceanTwoColumn\else
\begin{figure*}[t]
  \centering
  \includegraphics[width=\textwidth]{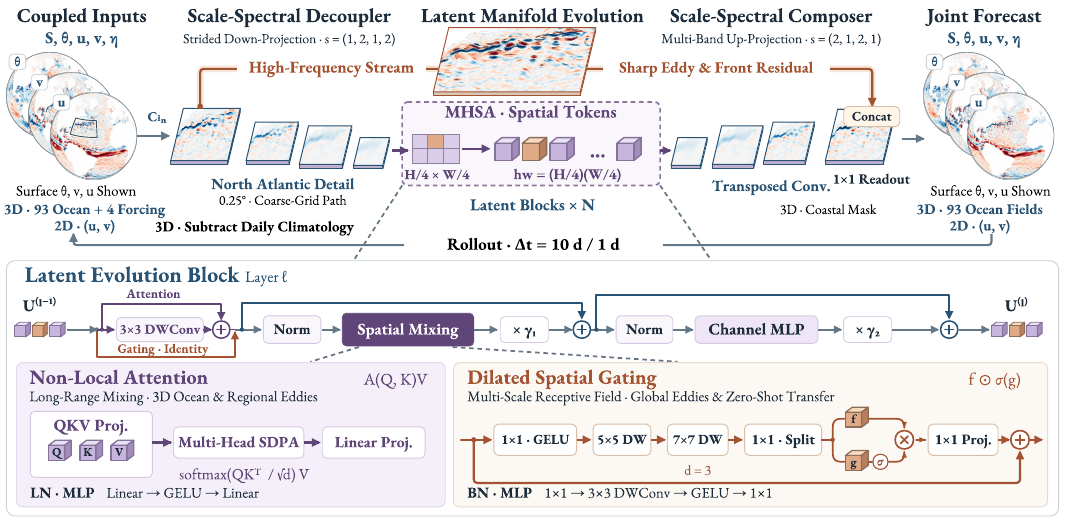}
  \caption{\textbf{\textit{\method architecture.}}
  The high-resolution stream carries fine-grid features to the decoder. On the coarse grid, the latent evolution block mixes spatial information through either non-local attention or resolution-invariant dilated spatial gating.}
  \label{fig:ocean_method}
\end{figure*}

\fi

This section presents \method, our scale-spectral decoupling framework (Fig.~\ref{fig:ocean_method}). Section~\ref{sec:method_problem} defines the unified prediction task and input representation. Section~\ref{sec:method_decoupler} introduces the scale-spectral decoupler, which separates the input into a coarse-grid evolution pathway and a full-resolution high-frequency stream. Section~\ref{sec:method_evolution} describes latent manifold evolution on the coarse grid and two spatial mixing operators. Section~\ref{sec:method_composer} presents the scale-spectral composer, and Section~\ref{sec:method_rollout} details closed-loop autoregressive rollout and multi-step unrolled training across tasks.

\subsection{Problem Formulation}
\label{sec:method_problem}

Let $\mathbf{X}_t\in\mathbb{R}^{C\times H\times W}$ denote the ocean state on day $t$ on a regular latitude-longitude grid, and let $\mathbf{F}_t\in\mathbb{R}^{C_f\times H\times W}$ denote the atmospheric surface forcing at the same time. Given the preceding $T$ days of ocean states and the following $T$ days of atmospheric surface forcing, the model predicts the subsequent $T$ days of ocean states through a conditional multi-step temporal block mapping given by
\begin{equation}
\begin{aligned}
\hat{\mathbf{X}}_{t+1:t+T}
&=\mathcal{F}_\theta\!\left(
\left[\mathbf{X}_{t-T+1:t},\,\mathbf{F}_{t+1:t+T}\right]
\right).
\end{aligned}
\label{eq:method_prediction}
\end{equation}
Here, $[\cdot\,,\cdot]$ denotes framewise concatenation along the channel dimension. We denote the input tensor by $\mathbf{S}\in\mathbb{R}^{T\times(C+C_f)\times H\times W}$ and the predicted output tensor by $\hat{\mathbf{S}}\in\mathbb{R}^{T\times C\times H\times W}$ across both ocean modeling regimes.

The block length $T$ and forcing channel count $C_f$ unify the two dynamical regimes considered in this work.
\begin{itemize}
\item[\ding{182}] \textbf{Surface eddy prediction.} The state consists of surface geostrophic velocities $(u,v)$ at $0.25^\circ$ (global) or $0.125^\circ$ (regional) resolution, with $C=2$, $T=10$, and no external atmospheric surface forcing channels ($C_f=0$).
\item[\ding{183}] \textbf{Three-dimensional ocean prediction.} The state contains temperature, salinity, and horizontal velocities $(u,v)$ across 23 depth levels together with sea surface height on a global $0.25^\circ$ grid, giving $C=93$ channels and $T=1$. Surface forcing comprises 10~m wind components, 2~m air temperature, and mean sea level pressure ($C_f=4$).
\end{itemize}

\noindent\textbf{\textit{Climatological Anomaly Representation.}}
For three-dimensional ocean prediction, we operate in a climatological anomaly space. Because the seasonal cycle accounts for a dominant fraction of raw ocean variance, predicting full fields directly biases network capacity toward stationary seasonal patterns instead of transient circulation anomalies. We subtract the daily climatology $\bar{\mathbf{X}}(d_t)$ for day-of-year $d_t$ and standardize each state channel using training statistics
\begin{equation}
\tilde{\mathbf{X}}_t
=\frac{\mathbf{X}_t-\bar{\mathbf{X}}(d_t)-\boldsymbol{\mu}}
{\boldsymbol{\sigma}},
\label{eq:method_anomaly}
\end{equation}
where $\boldsymbol{\mu}$ and $\boldsymbol{\sigma}$ are the channelwise training-set means and standard deviations, and land cells are masked to zero via the land-sea mask $\mathbf{M}\in\{0,1\}^{C\times H\times W}$. Subsequent inputs and outputs for the 3D task lie in this normalized anomaly space.

\subsection{Scale-Spectral Decoupler}
\label{sec:method_decoupler}

The decoupler processes each input frame independently through four convolutional units, each consisting of a $3\times3$ convolution, group normalization, and LeakyReLU
\begin{equation}
\begin{aligned}
\Phi_\ell(\mathbf{x})
&=\mathrm{LeakyReLU}\!\left(
\mathrm{GN}\!\left(\mathrm{Conv}_{3\times3}^{\,s_\ell}(\mathbf{x})\right)
\right),\\
(s_1,s_2,s_3,s_4)&=(1,2,1,2),
\end{aligned}
\label{eq:method_encoder_unit}
\end{equation}
where $s_\ell$ is the spatial stride. For each input frame $\mathbf{S}_i$, the decoupler splits the representation into two orthogonal pathways across spatial resolution and wavenumber via
\begin{equation}
\begin{aligned}
\mathbf{E}_i&=\Phi_1(\mathbf{S}_i)
\in\mathbb{R}^{D\times H\times W},\\
\mathbf{Z}_i&=(\Phi_4\circ\Phi_3\circ\Phi_2)(\mathbf{E}_i)
\in\mathbb{R}^{D\times\frac{H}{4}\times\frac{W}{4}}.
\end{aligned}
\label{eq:method_split}
\end{equation}

\noindent\textbf{\textit{Coarse-Grid Pathway.}}
Two stride-2 down-projections map $\mathbf{Z}_i$ to one quarter of the native resolution along each spatial axis. This projection acts as a learned low-pass filter that concentrates large-scale background structure into a compact latent manifold for deep temporal evolution~\cite{wu2024nmo}, simultaneously lowering computational cost and isolating the evolution core from high-frequency grid-scale phase noise.

\noindent\textbf{\textit{High-Frequency Stream.}}
Extracted at full resolution by the stride-1 unit $\Phi_1$, $\mathbf{E}_i$ captures sharp fronts and mesoscale eddies spanning only a few grid cells. Instead of adding intermediate skip connections across all downsampling stages, which would reintroduce high-frequency perturbations into the latent evolution path, we route only $\mathbf{E}_i$ around the entire evolution bottleneck to the final stage of the composer (Section~\ref{sec:method_composer}). This keeps the deep evolution core isolated on the coarse manifold while preserving fine-scale structures~\cite{li2025frequency}.

\subsection{Latent Manifold Evolution}
\label{sec:method_evolution}

\noindent\textbf{\textit{Temporal Folding.}}
We concatenate the $T$ coarse-grid latent frames along the channel axis to form a joint spatiotemporal latent tensor across the entire prediction window via
\begin{equation}
\begin{aligned}
\mathbf{U}^{(0)}&=[\mathbf{Z}_1,\dots,\mathbf{Z}_T]
\in\mathbb{R}^{TD\times h\times w},\\
h&=\tfrac{H}{4},\qquad w=\tfrac{W}{4}.
\end{aligned}
\label{eq:method_temporal_folding}
\end{equation}
Under this folding, channel projections jointly couple the $T$ time steps, allowing the evolution core to advance the entire temporal block in a single forward pass.

\noindent\textbf{\textit{Residual Evolution Blocks.}}
The latent state evolves through $N$ sequential residual blocks ($\ell=1,\dots,N$) given by
\begin{equation}
\mathbf{U}^{(\ell)}
=\mathcal{P}_\ell\!\left(
\mathcal{B}_\ell\!\left(\mathbf{U}^{(\ell-1)}\right)\right).
\label{eq:method_blocks}
\end{equation}
Here, $\mathcal{P}_1$ and $\mathcal{P}_N$ are $1\times1$ convolutions that map the channel width from $TD$ to hidden dimension $D_h$ and back to $TD$, while $\mathcal{P}_\ell$ is the identity for $1<\ell<N$. Each block $\mathcal{B}$ alternates spatial mixing $\mathcal{A}$ and channel mixing $\mathrm{MLP}$, modulated by learnable channelwise scaling vectors $\boldsymbol{\gamma}_1$ and $\boldsymbol{\gamma}_2$ via
\begin{equation}
\begin{aligned}
\mathbf{U}'&=\mathbf{U}+\boldsymbol{\gamma}_1\odot
\mathcal{A}\!\left(\mathrm{Norm}(\mathbf{U})\right),\\
\mathcal{B}(\mathbf{U})&=\mathbf{U}'+\boldsymbol{\gamma}_2\odot
\mathrm{MLP}\!\left(\mathrm{Norm}(\mathbf{U}')\right).
\end{aligned}
\label{eq:method_residual}
\end{equation}
Initializing $\boldsymbol{\gamma}_1$ and $\boldsymbol{\gamma}_2$ near zero keeps each block close to an identity mapping at the start of training, stabilizing optimization across deep rollouts. We instantiate the spatial mixing operator $\mathcal{A}$ in two forms according to whether the task requires cross-basin coupling on a fixed grid or resolution-invariant local operators for zero-shot transfer.

\noindent\textbf{\textit{Non-Local Attention.}}
Three-dimensional baroclinic circulation and energetic boundary currents exhibit long-range spatial coupling driven by Rossby wave propagation and basin-scale pressure adjustments. To capture these non-local interactions, we instantiate $\mathcal{A}$ with multi-head self-attention. Prior to normalization in each block, a $3\times3$ depthwise convolution injects local relative positional structure via
\begin{equation}
\mathbf{U}\leftarrow\mathbf{U}
+\mathrm{DWConv}_{3\times3}(\mathbf{U}).
\label{eq:method_position}
\end{equation}
Flattening the $h\times w$ latent grid into $hw$ spatial tokens yields
\begin{equation}
\begin{aligned}
\mathcal{A}_{\mathrm{attn}}(\mathbf{U})
&=\mathrm{softmax}\!\left(
\frac{\mathbf{Q}\mathbf{K}^{\top}}{\sqrt{d}}\right)
\mathbf{V}\mathbf{W}_o,\\
\mathbf{Q}&=\mathbf{U}\mathbf{W}_q,\quad
\mathbf{K}=\mathbf{U}\mathbf{W}_k,\quad
\mathbf{V}=\mathbf{U}\mathbf{W}_v.
\end{aligned}
\label{eq:method_attention}
\end{equation}
Here, $\mathrm{Norm}$ is layer normalization, and $\mathrm{MLP}$ is a two-layer feed-forward network with GELU activation. Operating on the $h\times w$ coarse grid reduces the token count by a factor of $16$, making global attention tractable across high-resolution ocean grids and multi-layer 3D circulation.

\noindent\textbf{\textit{Dilated Spatial Gating.}}
For global surface eddy fields and zero-shot cross-resolution transfer, the evolution operator must remain strictly invariant to grid dimensions. We design a purely convolutional multiscale gating operator given by
\begin{equation}
\begin{aligned}
\mathcal{A}_{\mathrm{gate}}(\mathbf{U})
&=\mathbf{U}+\mathbf{W}_2\,
\mathcal{G}\!\left(\mathrm{GELU}(\mathbf{W}_1\mathbf{U})\right),\\
\mathcal{G}(\mathbf{Y})&=\mathbf{f}\odot\sigma(\mathbf{g}),\\
[\mathbf{f},\mathbf{g}]
&=\mathbf{W}_3\,
\mathrm{DWConv}_{7\times7}^{\,d=3}\!\left(
\mathrm{DWConv}_{5\times5}(\mathbf{Y})\right).
\end{aligned}
\label{eq:method_gating}
\end{equation}
Here, $\mathbf{W}_1$, $\mathbf{W}_2$, and $\mathbf{W}_3$ are $1\times1$ projections, $\mathrm{DWConv}^{\,d=3}$ denotes a dilation-3 depthwise convolution, and $\sigma$ is the sigmoid gate. Stacking a $5\times5$ local depthwise convolution with a $7\times7$ dilated depthwise convolution spans a $23\times23$ receptive field on the latent grid ($92\times92$ native grid cells), capturing both local strain and mesoscale eddy interactions. Multiplicative gating $\mathbf{f}\odot\sigma(\mathbf{g})$ selectively amplifies active filaments and eddy cores. In this instantiation, $\mathrm{Norm}$ is batch normalization, and $\mathrm{MLP}$ comprises a $1\times1$ convolution, a $3\times3$ depthwise convolution, GELU activation, and a final $1\times1$ projection without fixed positional embeddings.

Because every parameter in $\mathcal{A}_{\mathrm{gate}}$ belongs to a translation-equivariant kernel independent of $(H,W)$, a model trained on a $0.25^\circ$ grid transfers directly to a $0.125^\circ$ global ocean grid in a zero-shot cross-resolution setting.

\subsection{Scale-Spectral Composer}
\label{sec:method_composer}

After latent evolution, we unfold $\mathbf{U}^{(N)}$ along the channel dimension into $T$ evolved latent frames $\{\mathbf{Z}'_i\}_{i=1}^{T}$ and decode each frame in parallel. Mirroring the decoupler, the first three composer units apply spatial strides $(2,1,2)$ with transposed convolutions in the stride-2 stages to restore the native $H\times W$ spatial grid dimensions across all channels via
\begin{equation}
\mathbf{Y}_i=(\Psi_3\circ\Psi_2\circ\Psi_1)(\mathbf{Z}'_i)
\in\mathbb{R}^{D\times H\times W}.
\label{eq:method_upsampling}
\end{equation}
The final stride-1 unit $\Psi_4$ fuses the upsampled large-scale state $\mathbf{Y}_i$ with its corresponding full-resolution bypass feature $\mathbf{E}_i$, followed by a $1\times1$ readout projection $\mathbf{W}_{\mathrm{out}}$ and the binary land-sea mask $\mathbf{M}$ across all grid cells via
\begin{equation}
\hat{\mathbf{S}}_i
=\mathbf{M}\odot\mathbf{W}_{\mathrm{out}}\,
\Psi_4\!\left([\mathbf{Y}_i,\mathbf{E}_i]\right).
\label{eq:method_composition}
\end{equation}
In this composition, $\mathbf{Y}_i$ supplies the evolved low-wavenumber phase and background transport, while the $3\times3$ convolutional unit $\Psi_4$ uses this updated large-scale flow context to locally realign and refine the full-resolution frontal and eddy features in $\mathbf{E}_i$ instead of copying static input residuals, with $\mathbf{M}$ enforcing exact zero values over land grid cells across all autoregressive rollout steps.

\subsection{Autoregressive Rollout and Training}
\label{sec:method_rollout}

\noindent\textbf{\textit{Rollout.}}
At inference time, let $\tau_k=(t+(k-1)T+1,\dots,t+kT)$ denote the $k$-th $T$-day window, so \method rolls out recursively via
\begin{equation}
\hat{\mathbf{X}}_{\tau_{k+1}}
=\mathcal{F}_\theta\!\left(
\left[\hat{\mathbf{X}}_{\tau_k},\,\mathbf{F}_{\tau_{k+1}}\right]
\right),
\label{eq:method_rollout}
\end{equation}
with $\hat{\mathbf{X}}_{\tau_0}=\mathbf{X}_{\tau_0}$. For surface eddy prediction ($T=10$), a 120-day rollout requires only 12 autoregressive steps, substantially curbing iterative error accumulation. For three-dimensional ocean prediction ($T=1$), the model steps forward daily under atmospheric surface forcing from ERA5 reanalysis (in ocean-only simulation) or from a coupled atmospheric forecasting system across the prediction window.

\noindent\textbf{\textit{Training.}}
To account for meridian convergence on a regular latitude-longitude grid, we weight grid cells by the cosine of latitude across all zonal rows according to
\begin{equation}
w(\varphi_m)
=\frac{\cos\varphi_m}
{\frac{1}{H}\sum_{m'=1}^{H}\cos\varphi_{m'}},
\label{eq:method_latitude_weight}
\end{equation}
where $\varphi_m$ is the latitude of row $m$. We optimize the latitude-weighted masked error over valid ocean cells given by
\begin{equation}
\begin{aligned}
\mathcal{L}
&=\frac{1}{T\sum\mathbf{M}}
\sum_{i=1}^{T}\sum_{c=1}^{C}\sum_{m=1}^{H}\sum_{n=1}^{W}
w(\varphi_m)\,M_{c,m,n}\\
&\hspace{5em}\cdot
\left(\hat{S}_{i,c,m,n}-X_{t+i,c,m,n}\right)^2,
\end{aligned}
\label{eq:method_loss}
\end{equation}
where $\sum\mathbf{M}$ denotes the total number of valid ocean entries per frame. Following single-step pretraining, we perform $K$-step unrolled fine-tuning by supervising $K$ consecutive rollout steps with Eq.~\eqref{eq:method_loss}, aligning training dynamics with closed-loop autoregressive evaluation across horizons.

\section{Experiments}
\label{sec:experiments}

Our experiments address four research questions aligned with our method.
\mbox{\ding{182}\,\textbf{\textit{RQ1}}}~How does \method perform against existing ocean foundation models, spatiotemporal backbones, and neural operators in global 3D ocean prediction and 120-day mesoscale eddy forecasting?
\mbox{\ding{183}\,\textbf{\textit{RQ2}}}~Does scale-spectral decoupling mitigate high-wavenumber energy dissipation and preserve spectral structure across regional and global rollouts?
\mbox{\ding{184}\,\textbf{\textit{RQ3}}}~Can a global $0.25^\circ$ model equipped with dilated spatial gating transfer directly to a $0.125^\circ$ grid in a zero-shot setting?
\mbox{\ding{185}\,\textbf{\textit{RQ4}}}~How do coarse-grid latent evolution, the single full-resolution bypass, and the spatial mixing operators individually contribute to resolving the spectral dilemma across long-horizon rollouts?

\subsection{Experimental Setup}
\label{sec:experimental_setup}

\begingroup
\definecolor{resultink}{HTML}{1E2235}
\definecolor{resultpurple}{HTML}{4A3574}
\definecolor{resultheader}{HTML}{E6E4F2}
\definecolor{resultstripe}{HTML}{F5F6FA}
\definecolor{resultours}{HTML}{ECE6F8}
\definecolor{rankfirst}{HTML}{DDD0F2}
\definecolor{ranksecond}{HTML}{F8E5D0}
\definecolor{rankthird}{HTML}{DCEBF5}
\definecolor{resultpending}{HTML}{8A8FA0}
\definecolor{gainfg}{HTML}{1C7254}

\newcommand{\rankone}[1]{{\setlength{\fboxsep}{0pt}\cellcolor{rankfirst}}\textbf{#1}}
\newcommand{\ranktwo}[1]{{\setlength{\fboxsep}{0pt}\cellcolor{ranksecond}}#1}
\newcommand{\rankthree}[1]{{\setlength{\fboxsep}{0pt}\cellcolor{rankthird}}#1}
\newcommand{\naentry}{\textcolor{resultpending}{N/A}}
\newcommand{\colmetric}[1]{{\fontsize{6.9}{8.1}\selectfont\mbox{#1}}}
\newcommand{\gainval}[1]{{\fontsize{6.8}{8.0}\selectfont\textcolor{gainfg}{\bfseries\mbox{#1}}}}
\newcommand{\lossval}[1]{{\fontsize{6.8}{8.0}\selectfont\textcolor{resultpending}{\mbox{#1}}}}

\begin{table*}[t]
\centering
\caption{\textbf{\textit{Global 3D ocean simulation and coupled forecasting.}}
Latitude-weighted RMSE ($\downarrow$) over 240 initial conditions in 2020 under ERA5 forcing and coupled atmospheric forecasting across four surface variables, rounded to two significant figures with ranks from full precision.}
\vspace{-3pt}
\label{tab:ocean3d_main}
\fontencoding{T1}\fontfamily{EBGaramond-TLF}\fontsize{7.8}{9.1}\selectfont
\color{resultink}
\setlength{\tabcolsep}{0pt}
\renewcommand{\arraystretch}{1.02}
\setlength{\aboverulesep}{1.0pt}
\setlength{\belowrulesep}{1.0pt}
\begin{tabularx}{\textwidth}{@{}p{0.128\textwidth}*{16}{>{\centering\arraybackslash}X}@{}}
\toprule[0.95pt]
\rowcolor{resultheader}
\rule{0pt}{10.4pt}\textbf{Method}
 & \multicolumn{4}{c}{\textbf{Temperature (${}^\circ\mathrm{C}$)$\downarrow$}}
 & \multicolumn{4}{c}{\textbf{Salinity ($\mathrm{PSU}$)$\downarrow$}}
 & \multicolumn{4}{c}{\textbf{Zonal $u$ ($\mathrm{m/s}$)$\downarrow$}}
 & \multicolumn{4}{c@{}}{\textbf{Meridional $v$ ($\mathrm{m/s}$)$\downarrow$}}\\[-0.5pt]
\cmidrule(lr){2-5}\cmidrule(lr){6-9}\cmidrule(lr){10-13}\cmidrule(l){14-17}
\rowcolor{resultheader}
 & \multicolumn{2}{c}{\colmetric{\textbf{Simulation}}} & \multicolumn{2}{c}{\colmetric{\textbf{Forecasting}}}
 & \multicolumn{2}{c}{\colmetric{\textbf{Simulation}}} & \multicolumn{2}{c}{\colmetric{\textbf{Forecasting}}}
 & \multicolumn{2}{c}{\colmetric{\textbf{Simulation}}} & \multicolumn{2}{c}{\colmetric{\textbf{Forecasting}}}
 & \multicolumn{2}{c}{\colmetric{\textbf{Simulation}}} & \multicolumn{2}{c@{}}{\colmetric{\textbf{Forecasting}}}\\[-0.5pt]
\cmidrule(lr){2-3}\cmidrule(lr){4-5}
\cmidrule(lr){6-7}\cmidrule(lr){8-9}
\cmidrule(lr){10-11}\cmidrule(lr){12-13}
\cmidrule(lr){14-15}\cmidrule(l){16-17}
 & \colmetric{Day 60} & \colmetric{Day 120} & \colmetric{Day 30} & \colmetric{Day 60}
 & \colmetric{Day 60} & \colmetric{Day 120} & \colmetric{Day 30} & \colmetric{Day 60}
 & \colmetric{Day 60} & \colmetric{Day 120} & \colmetric{Day 30} & \colmetric{Day 60}
 & \colmetric{Day 60} & \colmetric{Day 120} & \colmetric{Day 30} & \colmetric{Day 60}\\[-0.5pt]
\midrule[0.55pt]
\rowcolor{resultstripe} FourCastNet~\cite{pathak2022fourcastnet}
 & 1.6 & 6.4 & \rankthree{0.83} & 1.6
 & 1.0 & 2.7 & \ranktwo{0.41} & 1.0
 & 0.43 & 1.2 & \rankthree{0.19} & 0.42
 & 0.44 & 1.7 & \rankthree{0.17} & 0.43 \\
AI-GOMS~\cite{xiong2023ai}
 & 0.90 & 1.1 & \ranktwo{0.81} & \ranktwo{1.0}
 & 0.79 & 1.0 & \rankthree{0.53} & \rankthree{0.80}
 & \ranktwo{0.18} & \rankthree{0.22} & \ranktwo{0.17} & \ranktwo{0.20}
 & \rankthree{0.17} & \rankthree{0.20} & \ranktwo{0.16} & \ranktwo{0.19} \\
\rowcolor{resultstripe} ORCA-DL~\cite{guo2025orca}
 & \rankthree{0.89} & \ranktwo{1.0} & 0.98 & \rankthree{1.0}
 & \rankthree{0.49} & \ranktwo{0.54} & 0.61 & \ranktwo{0.64}
 & \rankthree{0.20} & \ranktwo{0.22} & 0.24 & \rankthree{0.25}
 & \ranktwo{0.17} & \rankone{0.18} & 0.19 & \rankthree{0.19} \\
WenHai~\cite{cui2025forecasting}
 & \ranktwo{0.87} & \rankthree{1.1} & \naentry & \naentry
 & \ranktwo{0.47} & \rankthree{0.62} & \naentry & \naentry
 & 0.21 & 0.25 & \naentry & \naentry
 & 0.18 & 0.21 & \naentry & \naentry \\
\addlinespace[1.4pt]
\midrule[0.6pt]
\rowcolor{resultours}
\rule{0pt}{11.0pt}\textcolor{resultpurple}{\bfseries\scshape QuantAtlas-OM}
 & \rankone{0.79} & \rankone{0.96} & \rankone{0.80} & \rankone{0.96}
 & \rankone{0.40} & \rankone{0.47} & \rankone{0.40} & \rankone{0.47}
 & \rankone{0.17} & \rankone{0.19} & \rankone{0.17} & \rankone{0.20}
 & \rankone{0.17} & \ranktwo{0.18} & \rankone{0.16} & \rankone{0.19}\\[1pt]
\addlinespace[1.1pt]
{\fontsize{7.3}{8.6}\selectfont\textcolor{gainfg}{\itshape $\Delta$ vs.\ best}}
 & \gainval{$-$9.3\%} & \gainval{$-$4.4\%} & \gainval{$-$1.3\%} & \gainval{$-$5.8\%}
 & \gainval{$-$15\%} & \gainval{$-$13\%} & \gainval{$-$3.1\%} & \gainval{$-$27\%}
 & \gainval{$-$7.2\%} & \gainval{$-$12\%} & \gainval{$-$2.2\%} & \gainval{$-$2.7\%}
 & \gainval{$-$2.1\%} & \lossval{+4.2\%} & \gainval{$-$0.80\%} & \gainval{$-$1.3\%}\\[0.5pt]
\bottomrule[0.95pt]
\end{tabularx}
\vspace{-4pt}
\end{table*}
\endgroup

\begingroup
\definecolor{resultink}{HTML}{1E2235}
\definecolor{resultpurple}{HTML}{4A3574}
\definecolor{resultheader}{HTML}{E6E4F2}
\definecolor{resultstripe}{HTML}{F5F6FA}
\definecolor{resultours}{HTML}{ECE6F8}
\definecolor{rankfirst}{HTML}{DDD0F2}
\definecolor{ranksecond}{HTML}{F8E5D0}
\definecolor{rankthird}{HTML}{DCEBF5}
\definecolor{resultpending}{HTML}{8A8FA0}
\definecolor{gainfg}{HTML}{1C7254}

\newcommand{\rankone}[1]{{\setlength{\fboxsep}{0pt}\cellcolor{rankfirst}}\textbf{#1}}
\newcommand{\ranktwo}[1]{{\setlength{\fboxsep}{0pt}\cellcolor{ranksecond}}#1}
\newcommand{\rankthree}[1]{{\setlength{\fboxsep}{0pt}\cellcolor{rankthird}}#1}
\newcommand{\naentry}{\textcolor{resultpending}{\textendash}}
\newcommand{\resultcategory}[1]{%
  \addlinespace[1.5pt]
  \multicolumn{20}{@{}l}{\rule{0pt}{9.8pt}\textcolor{resultpurple}{\bfseries\itshape $\blacktriangleright$~#1}}\\[0.4pt]}
\newcommand{\colmetric}[1]{{\fontsize{6.6}{7.8}\selectfont\mbox{#1}}}
\newcommand{\gainval}[1]{{\fontsize{6.2}{7.4}\selectfont\textcolor{gainfg}{\bfseries\mbox{#1}}}}

\begin{table*}[t]
\centering
\caption{\textbf{\textit{Regional mesoscale eddy forecasting.}}
Latitude-weighted RMSE ($\mathrm{m/s}$, $\downarrow$) and ACC ($\uparrow$) averaged over $(u,v)$ across 30 initial conditions in 2023 on a unified $0.25^\circ$ scoring grid at 10, 60, and 120 days, rounded to two significant figures with ranks from full precision.}
\vspace{-3pt}
\label{tab:regional_eddy_main}
\fontencoding{T1}\fontfamily{EBGaramond-TLF}\fontsize{7.3}{8.6}\selectfont
\color{resultink}
\setlength{\tabcolsep}{0pt}
\renewcommand{\arraystretch}{0.94}
\setlength{\aboverulesep}{1.0pt}
\setlength{\belowrulesep}{1.0pt}
\begin{tabularx}{\textwidth}{@{}p{0.116\textwidth}>{\centering\arraybackslash}p{0.064\textwidth}*{18}{>{\centering\arraybackslash}X}@{}}
\toprule[0.95pt]
\rowcolor{resultheader}
\rule{0pt}{10.5pt}\textbf{Method}
 & \textbf{Family}
 & \multicolumn{6}{c}{\textbf{Kuroshio Extension}}
 & \multicolumn{6}{c}{\textbf{Gulf Stream}}
 & \multicolumn{6}{c@{}}{\textbf{Agulhas Current}}\\[-0.5pt]
\cmidrule(lr){3-8}\cmidrule(lr){9-14}\cmidrule(l){15-20}
\rowcolor{resultheader}
 &
 & \multicolumn{2}{c}{\colmetric{\textbf{Day 10}}} & \multicolumn{2}{c}{\colmetric{\textbf{Day 60}}} & \multicolumn{2}{c}{\colmetric{\textbf{Day 120}}}
 & \multicolumn{2}{c}{\colmetric{\textbf{Day 10}}} & \multicolumn{2}{c}{\colmetric{\textbf{Day 60}}} & \multicolumn{2}{c}{\colmetric{\textbf{Day 120}}}
 & \multicolumn{2}{c}{\colmetric{\textbf{Day 10}}} & \multicolumn{2}{c}{\colmetric{\textbf{Day 60}}} & \multicolumn{2}{c@{}}{\colmetric{\textbf{Day 120}}}\\[-0.5pt]
\cmidrule(lr){3-4}\cmidrule(lr){5-6}\cmidrule(lr){7-8}
\cmidrule(lr){9-10}\cmidrule(lr){11-12}\cmidrule(lr){13-14}
\cmidrule(lr){15-16}\cmidrule(lr){17-18}\cmidrule(l){19-20}
 & & \colmetric{RMSE$\downarrow$} & \colmetric{ACC$\uparrow$} & \colmetric{RMSE$\downarrow$} & \colmetric{ACC$\uparrow$} & \colmetric{RMSE$\downarrow$} & \colmetric{ACC$\uparrow$}
 & \colmetric{RMSE$\downarrow$} & \colmetric{ACC$\uparrow$} & \colmetric{RMSE$\downarrow$} & \colmetric{ACC$\uparrow$} & \colmetric{RMSE$\downarrow$} & \colmetric{ACC$\uparrow$}
 & \colmetric{RMSE$\downarrow$} & \colmetric{ACC$\uparrow$} & \colmetric{RMSE$\downarrow$} & \colmetric{ACC$\uparrow$} & \colmetric{RMSE$\downarrow$} & \colmetric{ACC$\uparrow$}\\[-0.5pt]
\midrule[0.55pt]
Persistence & \textit{Static} & 0.16 & 0.69 & 0.24 & 0.30 & 0.25 & \ranktwo{0.26} & 0.20 & 0.52 & 0.29 & 0.054 & 0.29 & 0.029 & 0.25 & 0.50 & 0.34 & 0.047 & 0.35 & {\fontsize{6.9}{8.2}\selectfont \textminus{}0.014} \\
\resultcategory{Spatiotemporal Backbones}
\rowcolor{resultstripe} U-Net & \textit{Conv} & 0.098 & 0.88 & \ranktwo{0.19} & 0.39 & \ranktwo{0.21} & 0.20 & 0.12 & 0.80 & \ranktwo{0.21} & 0.17 & \ranktwo{0.22} & 0.080 & 0.15 & 0.80 & \ranktwo{0.26} & 0.11 & \ranktwo{0.26} & 0.0069 \\
ResNet & \textit{Conv} & \ranktwo{0.090} & \ranktwo{0.90} & 0.40 & 0.26 & 380 & 0.068 & 0.12 & 0.84 & 0.25 & 0.13 & 0.31 & 0.035 & 0.14 & 0.83 & 0.28 & 0.099 & 0.31 & 0.028 \\
\rowcolor{resultstripe} ConvLSTM & \textit{RNN} & 0.10 & 0.87 & 0.19 & 0.37 & 0.72 & 0.056 & 0.13 & 0.77 & 0.75 & 0.022 & 1.7 & 0.010 & 0.16 & 0.77 & 1.1 & {\fontsize{6.9}{8.2}\selectfont \textminus{}0.0088} & 2.5 & 0.0037 \\
SimVP~\cite{gao2022simvp} & \textit{Conv} & \rankthree{0.091} & \rankthree{0.90} & 0.28 & 0.14 & 0.28 & 0.084 & \rankthree{0.11} & \rankthree{0.84} & 0.30 & 0.090 & 0.32 & 0.039 & \rankthree{0.14} & \rankthree{0.84} & 0.32 & 0.069 & 0.34 & {\fontsize{6.9}{8.2}\selectfont \textminus{}0.0079} \\
\rowcolor{resultstripe} PastNet~\cite{wu2024pastnet} & \textit{Conv} & 0.13 & 0.75 & 0.22 & 0.31 & 0.24 & 0.16 & 0.13 & 0.77 & 0.23 & \rankthree{0.19} & 0.26 & 0.031 & 0.15 & 0.78 & 0.29 & 0.13 & 0.30 & \rankthree{0.035} \\
{\fontsize{7.0}{8.4}\selectfont\mbox{Earthformer~\cite{gao2022earthformer}}} & \textit{Attn} & 0.097 & 0.88 & 0.21 & 0.40 & 0.23 & \rankthree{0.23} & 0.12 & 0.81 & 0.22 & 0.11 & 0.25 & {\fontsize{6.9}{8.2}\selectfont \textminus{}0.050} & 0.14 & 0.83 & 0.28 & 0.13 & 0.29 & 0.019 \\
\resultcategory{Neural PDE Operators}
\rowcolor{resultstripe} FNO~\cite{li2021fourier} & \textit{Spectral} & 0.11 & 0.85 & \rankthree{0.19} & \rankthree{0.40} & \rankthree{0.22} & 0.22 & 0.13 & 0.77 & 0.22 & \ranktwo{0.21} & 0.23 & \rankthree{0.11} & 0.16 & 0.77 & \rankthree{0.26} & 0.12 & \rankthree{0.27} & 0.022 \\
CNO~\cite{raonic2023convolutional} & \textit{Operator} & 0.091 & 0.89 & 0.21 & 0.36 & 0.25 & 0.12 & \ranktwo{0.11} & \ranktwo{0.85} & 0.22 & 0.18 & 0.23 & \ranktwo{0.12} & 0.14 & 0.84 & 0.28 & 0.076 & 0.29 & 0.0067 \\
\rowcolor{resultstripe} LSM & \textit{Spectral} & 0.094 & 0.89 & 0.19 & \ranktwo{0.43} & 0.23 & 0.21 & 0.11 & 0.84 & \rankthree{0.22} & 0.19 & \rankthree{0.23} & 0.082 & \ranktwo{0.13} & \ranktwo{0.84} & 0.28 & \ranktwo{0.15} & 0.30 & 0.0011 \\
KNO & \textit{Operator} & 0.11 & 0.83 & 0.22 & 0.29 & 1.0 & 0.045 & 0.14 & 0.75 & 0.24 & 0.091 & 0.32 & 0.033 & 0.17 & 0.74 & 0.29 & 0.067 & 0.53 & 0.0013 \\
\rowcolor{resultstripe} MgNO & \textit{Operator} & 0.097 & 0.88 & 0.20 & 0.34 & 0.23 & 0.18 & 0.12 & 0.81 & 0.22 & 0.16 & 0.23 & 0.094 & 0.14 & 0.82 & 0.27 & \rankthree{0.14} & 0.28 & \ranktwo{0.046} \\
\addlinespace[1.4pt]
\midrule[0.6pt]
\rowcolor{resultours}
\rule{0pt}{11.0pt}{\fontsize{7.0}{8.4}\selectfont\textcolor{resultpurple}{\bfseries\scshape\mbox{QuantAtlas-OM}}}
 & {\fontsize{6.8}{8.2}\selectfont\textcolor{gainfg}{\bfseries\itshape\mbox{Decoupled}}}
 & \rankone{0.016} & \rankone{0.99} & \rankone{0.084} & \rankone{0.88} & \rankone{0.15} & \rankone{0.66} & \rankone{0.020} & \rankone{0.99} & \rankone{0.11} & \rankone{0.81} & \rankone{0.21} & \rankone{0.39} & \rankone{0.019} & \rankone{0.99} & \rankone{0.11} & \rankone{0.84} & \rankone{0.25} & \rankone{0.35}\\[1pt]
\addlinespace[1.1pt]
{\fontsize{7.1}{8.4}\selectfont\textcolor{gainfg}{\itshape $\Delta$ vs.\ best}}
 & \naentry
 & \gainval{$-$82\%} & \gainval{+0.096} & \gainval{$-$55\%} & \gainval{2.1$\times$} & \gainval{$-$25\%} & \gainval{2.5$\times$} & \gainval{$-$82\%} & \gainval{+0.14} & \gainval{$-$48\%} & \gainval{3.8$\times$} & \gainval{$-$4.6\%} & \gainval{3.1$\times$} & \gainval{$-$86\%} & \gainval{+0.15} & \gainval{$-$58\%} & \gainval{5.7$\times$} & \gainval{$-$3.6\%} & \gainval{7.7$\times$}\\[0.5pt]
\bottomrule[0.95pt]
\end{tabularx}
\vspace{2pt}
\end{table*}
\endgroup

\begin{figure*}[t]
\centering
\includegraphics[width=\textwidth]{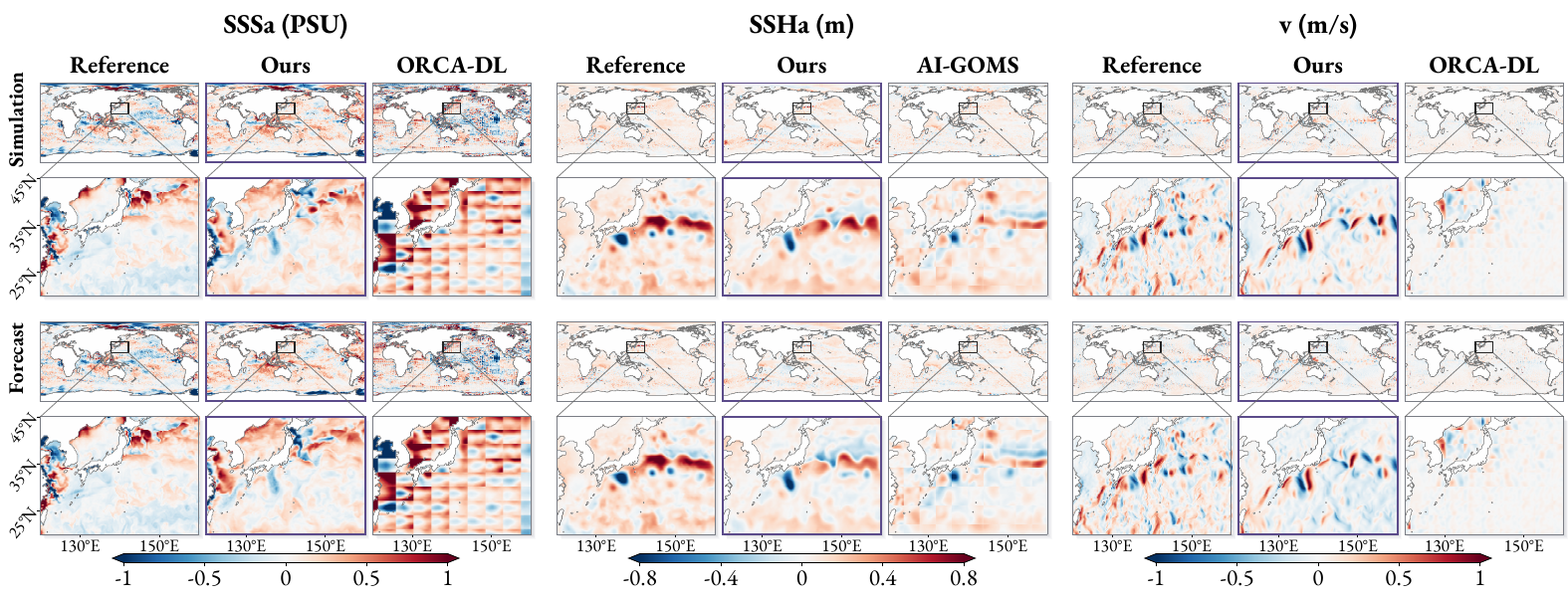}
\vspace{-6pt}
\caption{\textbf{\textit{Global 3D ocean surface fields and Kuroshio zooms at Day 60.}}
Columns compare the reference, \method (Ours), and the strongest baseline for sea surface salinity anomaly (SSSa), sea surface height anomaly (SSHa), and meridional velocity $v$ under ERA5-forced simulation (top) and coupled atmospheric forecasting (bottom) across global and local views.}
\label{fig:ocean3d_day60}
\vspace{0pt}
\includegraphics[width=\textwidth]{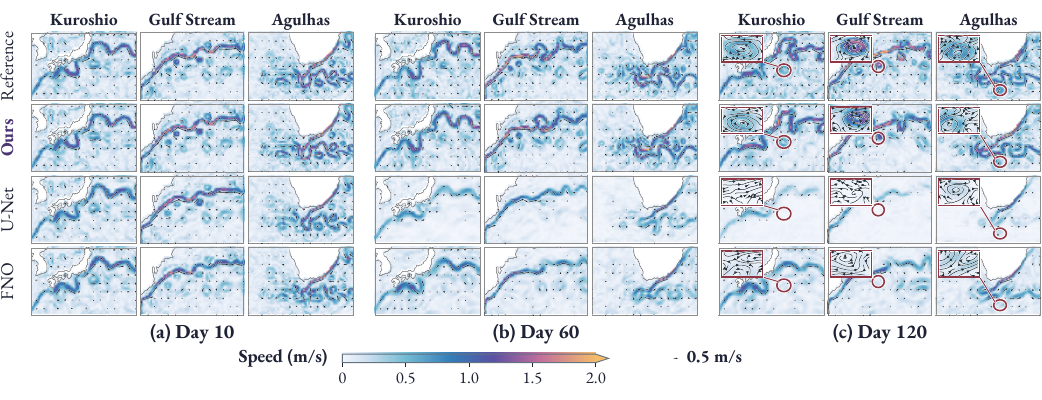}
\vspace{-6pt}
\caption{\textbf{\textit{Multi-horizon mesoscale eddy forecasts across three boundary currents.}}
Surface geostrophic speed $\sqrt{u^2+v^2}$ ($\mathrm{m/s}$) and velocity vectors at Days~10, 60, and 120 across the Kuroshio, Gulf Stream, and Agulhas Current, where red ellipses and Kuroshio streamline insets highlight preserved Day-120 mesoscale eddies, closed rotational cores, and sharp frontal jets across all three regions.}
\vspace{4pt}
\label{fig:three_leads_speed}
\end{figure*}

\noindent\textbf{\textit{Tasks, Datasets, and Protocol.}}
We evaluate \method across four settings covering global $0.25^\circ$ 3D ocean simulation ($23$ levels, $93$ state channels, and $4$ ERA5 forcing channels on a $720\times1440$ grid from GLORYS12~\cite{lellouche2021glorys}), global $0.25^\circ$ coupled ocean-atmosphere forecasting, regional $0.125^\circ$ mesoscale eddy forecasting across the Kuroshio Extension, Gulf Stream, and Agulhas Current using CMEMS geostrophic velocities (\texttt{ugos}, \texttt{vgos}), and zero-shot $0.25^\circ\to0.125^\circ$ global transfer. Trained on $1993$ to $2017$ ($3$D ocean) and $1993$ to $2018$ (eddies) records, closed-loop rollouts without future ocean observations span $120$-day ERA5-forced simulations and $60$-day coupled forecasts over $240$ initial conditions in $2020$ for the $3$D ocean, twelve $10$-day blocks ($120$ days) over $30$ initial conditions in $2023$ on a unified $0.25^\circ$ scoring grid for regional eddies, and a $10$-day native $0.125^\circ$ ($1440\times2880$) rollout in $2020$ for zero-shot global surface-current transfer.

\noindent\textbf{\textit{Baselines, Metrics, and Implementation.}}
Baselines include regional Persistence (last observed state) and \mbox{\ding{70}}~ocean and Earth system models (FourCastNet~\cite{pathak2022fourcastnet}, AI-GOMS~\cite{xiong2023ai}, ORCA-DL~\cite{guo2025orca}, WenHai~\cite{cui2025forecasting}), \mbox{\ding{70}}~convolutional and recurrent backbones (U-Net, ResNet, ConvLSTM, SimVP~\cite{gao2022simvp}, PastNet~\cite{wu2024pastnet}), \mbox{\ding{70}}~spatiotemporal Transformers (Earthformer~\cite{gao2022earthformer}), and \mbox{\ding{70}}~neural PDE operators (FNO~\cite{li2021fourier}, CNO~\cite{raonic2023convolutional}, LSM, KNO, MgNO). We report latitude-weighted RMSE, ACC, spatial PCC, radially averaged kinetic energy spectra $E(k)$, and two-dimensional angular-grid power spectra across all forecast horizons.

\subsection{Global and Mesoscale Forecasting (RQ1)}
\label{sec:exp_rq1}

Table~\ref{tab:ocean3d_main} and Fig.~\ref{fig:ocean3d_day60} evaluate global $0.25^\circ$, $23$-level three-dimensional ocean simulations over $120$ days and coupled ocean-atmosphere forecasts over $60$ days. Whereas FourCastNet diverges at long horizons, \method ranks first on $15$ of $16$ metrics across surface temperature, salinity, and horizontal velocities $u$ and $v$, cutting $60$- to $120$-day salinity RMSE by $13\%$ to $27\%$ over the second-best model. In the Day-$60$ global fields and Kuroshio zooms of Fig.~\ref{fig:ocean3d_day60}, ORCA-DL exhibits blocky salinity artifacts and loses meridional velocity streaks, and AI-GOMS over-smooths sea surface height eddy dipoles, whereas \method faithfully reconstructs sharp salinity fronts, eddy trains, and alternating velocity bands under both simulation and coupled forecasting. Table~\ref{tab:regional_eddy_main} and Fig.~\ref{fig:three_leads_speed} further demonstrate this advantage in $120$-day autoregressive mesoscale eddy forecasting across the Kuroshio Extension, Gulf Stream, and Agulhas Current. At Day~$10$, \method lowers RMSE to $0.016$, $0.020$, and $0.019\,\mathrm{m/s}$ across the three domains, reducing error by $82\%$ to $86\%$ over the strongest baseline while maintaining $0.99$ ACC. By Day~$60$, when every competing method drops to $0.43$ ACC or below, \method sustains $0.81$ to $0.88$ ACC and cuts RMSE by $48\%$ to $58\%$ across all three boundary currents. At Day~$120$, while U-Net, FNO, and SimVP smooth vortices into weak parallel streamlines and ResNet and ConvLSTM diverge, \method achieves the lowest RMSE across all regions, reaches $0.66$ ACC in the Kuroshio ($2.5\times$ to $7.7\times$ the best baseline across regions), and preserves energetic frontal jets up to $2.0\,\mathrm{m/s}$ and closed rotational eddy cores across all three western boundary current systems in Fig.~\ref{fig:three_leads_speed}c without spectral blurring or long-horizon autoregressive phase drift across scales.

\subsection{Spectral Fidelity and Retention (RQ2)}
\label{sec:exp_rq2}

\begin{figure}[t!]
\centering
\includegraphics[width=\columnwidth]{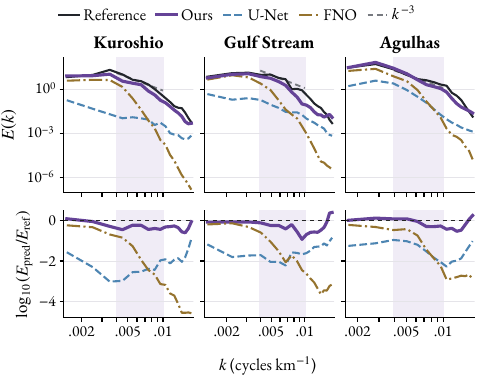}
\vspace{-6pt}
\caption{\textbf{\textit{Regional kinetic energy spectra at Day 120.}}
Radial spectra $E(k)$ (top) and $\log_{10}(E_{\mathrm{pred}}/E_{\mathrm{ref}})$ (bottom), with the $100$ to $250\,\mathrm{km}$ mesoscale band shaded and a $k^{-3}$ slope guide.}
\vspace{0pt}
\label{fig:regional_spectra}
\end{figure}

\begin{figure}[t!]
\centering
\includegraphics[width=\columnwidth]{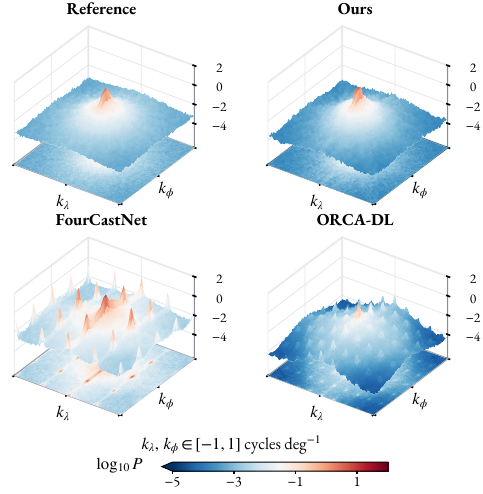}
\vspace{-6pt}
\caption{\textbf{\textit{Global surface-current 2D spectra at Day 60.}}
Angular-grid power spectra $\log_{10} P(k_\lambda, k_\phi)$ shown as 3D spectral surfaces with bottom projections under a shared color scale.}
\vspace{-3pt}
\label{fig:global_2d_spectra}
\end{figure}

Figures~\ref{fig:regional_spectra} and~\ref{fig:global_2d_spectra} examine spectral retention across long-horizon rollouts in the frequency domain. Figure~\ref{fig:regional_spectra} compares radial kinetic energy spectra $E(k)$ and logarithmic spectral ratios $\log_{10}(E_{\mathrm{pred}}/E_{\mathrm{ref}})$ at Day~$120$ across the three western boundary current systems from a shared initial condition. Within the $100$ to $250\,\mathrm{km}$ mesoscale band, U-Net and FNO exhibit pronounced high-wavenumber energy dissipation, as their band-integrated spectral energy ratios drop to $0.0023$ and $0.0790$ in the Kuroshio, $0.0129$ and $0.1069$ in the Gulf Stream, and $0.0584$ and $0.1957$ in the Agulhas Current, with $\log_{10}(E_{\mathrm{pred}}/E_{\mathrm{ref}})$ falling toward $-2$ to $-4$ at high wavenumbers. In contrast, \method retains band-integrated energy ratios of $0.4372$, $0.5744$, and $0.8649$ across the three regions, tracking the reference spectra and the $k^{-3}$ slope guide much more closely across scales. Figure~\ref{fig:global_2d_spectra} further visualizes 2D angular-grid power spectra $\log_{10} P(k_\lambda, k_\phi)$ of surface currents from a shared Day-$60$ coupled global 3D forecast as spectral height surfaces with floor projections. FourCastNet erupts into high-energy spectral ridges and spikes across discrete zonal wavenumbers, and ORCA-DL collapses the central energy peak while leaving periodic lattice bumps across the high-frequency plane matching its block artifacts in Fig.~\ref{fig:ocean3d_day60}, whereas \method closely matches both the central peak height and the radial decay envelope of the reference global ocean spectrum.

\subsection{Zero-Shot Resolution Transfer (RQ3)}
\label{sec:exp_rq3}

\begin{figure}[t!]
\centering
\includegraphics[width=\columnwidth]{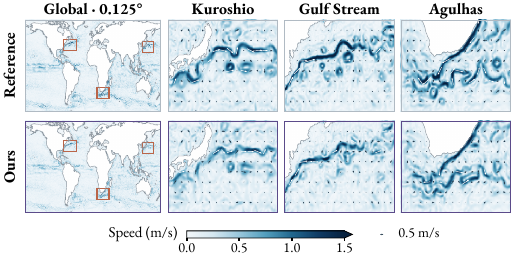}
\vspace{-6pt}
\caption{\textbf{\textit{Zero-shot $0.25^\circ\to0.125^\circ$ forecast at Day 10.}}
Direct inference on the $1440\times2880$ grid without fine-tuning, with three regional windows cropped from the same global prediction.}
\vspace{3pt}
\label{fig:zero_shot_day10}
\end{figure}

Figure~\ref{fig:zero_shot_day10} and Table~\ref{tab:zero_shot_day10} evaluate zero-shot cross-resolution transfer from the global $0.25^\circ$ ($720\times1440$) training grid to the native global $0.125^\circ$ ($1440\times2880$) grid without fine-tuning. Equipped with resolution-invariant dilated spatial gating ($\mathcal{A}_{\mathrm{gate}}$) and local convolutional operators in the latent space, the $0.25^\circ$-trained \method checkpoint directly executes a 10-day rollout on the $4\times$-denser $0.125^\circ$ ocean grid ($\sim\!2.70\times10^6$ valid ocean cells). As reported in Table~\ref{tab:zero_shot_day10},
\begingroup
\definecolor{resultink}{HTML}{1E2235}
\definecolor{resultpurple}{HTML}{4A3574}
\definecolor{resultheader}{HTML}{E6E4F2}
\definecolor{resultstripe}{HTML}{F5F6FA}
\definecolor{resultours}{HTML}{ECE6F8}
\begin{table}[t!]
\vspace{3pt}
\centering
\caption{\textbf{\textit{Zero-shot $0.25^\circ\to0.125^\circ$ metrics at Day 10.}}
Latitude-weighted RMSE, MAE ($\mathrm{m/s}$, $\downarrow$), and spatial PCC ($\uparrow$) evaluated directly on the native $1440\times2880$ global ocean grid.}
\vspace{-1pt}
\label{tab:zero_shot_day10}
\fontencoding{T1}\fontfamily{EBGaramond-TLF}\fontsize{7.9}{9.1}\selectfont
\color{resultink}
\setlength{\tabcolsep}{0pt}
\renewcommand{\arraystretch}{1.0}
\setlength{\aboverulesep}{0.9pt}
\setlength{\belowrulesep}{0.9pt}
\begin{tabularx}{\columnwidth}{@{}p{0.31\columnwidth}*{3}{>{\centering\arraybackslash}X}@{}}
\toprule[0.95pt]
\rowcolor{resultheader}
\rule{0pt}{9.2pt}\textbf{Variable}
 & \textbf{RMSE ($\mathrm{m/s}$)$\downarrow$}
 & \textbf{MAE ($\mathrm{m/s}$)$\downarrow$}
 & \textbf{Spatial PCC$\uparrow$}\\[-0.3pt]
\midrule[0.55pt]
\rowcolor{resultstripe} Zonal velocity $u$ & 0.0978 & 0.0532 & 0.8224 \\
Meridional velocity $v$ & 0.0826 & 0.0442 & 0.8036 \\
\rowcolor{resultours}
\rule{0pt}{9.4pt}\textcolor{resultpurple}{\bfseries $u/v$ Mean} & \textbf{0.0902} & \textbf{0.0487} & \textbf{0.8130} \\
\bottomrule[0.95pt]
\end{tabularx}
\vspace{-3pt}
\end{table}
\endgroup%
 at Day~$10$ the model maintains latitude-weighted RMSE and MAE of $0.0978$ and $0.0532\,\mathrm{m/s}$ for $u$ and $0.0826$ and $0.0442\,\mathrm{m/s}$ for $v$ ($0.0902\,\mathrm{m/s}$ RMSE and $0.0487\,\mathrm{m/s}$ MAE on average), while achieving a mean weighted spatial Pearson correlation coefficient (PCC) of $0.8130$ ($0.999$ at Day~$1$ to $0.971$ at Day~$5$). In the Kuroshio, Gulf Stream, and Agulhas windows cropped from the same global Day-$10$ prediction in Fig.~\ref{fig:zero_shot_day10}, the zero-shot $0.125^\circ$ rollout reconstructs meandering jet axes, sharp frontal gradients, and coherent rotational eddies without cross-grid spectral aliasing.

\subsection{Ablation and Spectral Dilemma (RQ4)}
\label{sec:exp_rq4}

\begingroup
\definecolor{resultink}{HTML}{1E2235}
\definecolor{resultpurple}{HTML}{4A3574}
\definecolor{resultheader}{HTML}{E6E4F2}
\definecolor{resultstripe}{HTML}{F5F6FA}
\definecolor{resultours}{HTML}{ECE6F8}
\definecolor{rankfirst}{HTML}{DDD0F2}
\definecolor{ranksecond}{HTML}{F8E5D0}

\newcommand{\rankone}[1]{{\setlength{\fboxsep}{0pt}\cellcolor{rankfirst}}\textbf{#1}}
\newcommand{\ranktwo}[1]{{\setlength{\fboxsep}{0pt}\cellcolor{ranksecond}}#1}
\newcommand{\colmetric}[1]{{\fontsize{6.6}{7.8}\selectfont\mbox{#1}}}

\begin{table}[t!]
\centering
\caption{\textbf{\textit{Ablation study on Kuroshio Extension ($120$-day rollout).}}
Latitude-weighted RMSE ($\mathrm{m/s}$, $\downarrow$), ACC ($\uparrow$), and Day-120 spectral ratio $R_{100\text{-}250}$ in the $100$ to $250\,\mathrm{km}$ mesoscale band.}
\vspace{-2pt}
\label{tab:ablation_kuroshio}
\fontencoding{T1}\fontfamily{EBGaramond-TLF}\fontsize{7.3}{8.5}\selectfont
\color{resultink}
\setlength{\tabcolsep}{0pt}
\renewcommand{\arraystretch}{0.99}
\setlength{\aboverulesep}{0.9pt}
\setlength{\belowrulesep}{0.9pt}
\begin{tabularx}{\columnwidth}{@{}p{0.32\columnwidth}*{4}{>{\centering\arraybackslash}X}>{\centering\arraybackslash}p{0.17\columnwidth}@{}}
\toprule[0.95pt]
\rowcolor{resultheader}
\rule{0pt}{9.2pt}\textbf{Variant}
 & \multicolumn{2}{c}{\textbf{Day 10}}
 & \multicolumn{2}{c}{\textbf{Day 120}}
 & \textbf{Spectrum}\\[-0.4pt]
\cmidrule(lr){2-3}\cmidrule(lr){4-5}\cmidrule(l){6-6}
\rowcolor{resultheader}
 & \colmetric{RMSE$\downarrow$} & \colmetric{ACC$\uparrow$}
 & \colmetric{RMSE$\downarrow$} & \colmetric{ACC$\uparrow$}
 & \colmetric{$R_{100\text{-}250}\uparrow$}\\[-0.4pt]
\midrule[0.55pt]
\rowcolor{resultstripe} w/o Outer bypass $\mathbf{E}_i$ & 0.042 & 0.96 & 0.19 & 0.42 & 0.094 \\
Multi-stage U-Net skips & 0.023 & 0.98 & 0.21 & 0.31 & 0.162 \\
\rowcolor{resultstripe} w/o Spatial mixing ($\mathcal{A}$) & 0.029 & 0.97 & 0.18 & 0.48 & 0.285 \\
Dilated gating ($\mathcal{A}_{\mathrm{gate}}$) & \ranktwo{0.019} & \ranktwo{0.99} & \ranktwo{0.16} & \ranktwo{0.61} & \ranktwo{0.391} \\
\addlinespace[0.8pt]
\midrule[0.6pt]
\rowcolor{resultours}
\rule{0pt}{9.4pt}\textcolor{resultpurple}{\bfseries\scshape\mbox{QuantAtlas-OM}}
 & \rankone{0.016} & \rankone{0.99}
 & \rankone{0.15} & \rankone{0.66}
 & \rankone{0.437} \\
\bottomrule[0.95pt]
\end{tabularx}
\vspace{0pt}
\end{table}
\endgroup

Table~\ref{tab:ablation_kuroshio} ablates \method on $120$-day Kuroshio Extension eddy forecasting to verify how each module resolves the spectral dilemma.
\mbox{\ding{182}\,\textbf{\textit{Bypass Topology.}}} Removing the outer full-resolution bypass (\textit{w/o Outer bypass $\mathbf{E}_i$}) forces fine eddies through four downsampling stages, so Day-$10$ RMSE degrades from $0.016$ to $0.042\,\mathrm{m/s}$ and the Day-$120$ mesoscale spectral energy ratio $R_{100\text{-}250}$ collapses from $0.437$ to $0.094$. Conversely, adding dense intermediate skips (\textit{Multi-stage U-Net skips}) preserves Day-$10$ detail ($0.023\,\mathrm{m/s}$) but injects high-frequency errors into the latent processor, dropping Day-$120$ ACC to $0.31$. Routing high-frequency structure solely through the single outer bypass $\mathbf{E}_i$ and composer $\Psi_4$ secures both Day-$120$ stability ($0.66$ ACC) and spectral retention ($0.437$).
\mbox{\ding{183}\,\textbf{\textit{Latent Mixing Operators.}}} Removing global spatial mixing in the latent blocks (\textit{w/o Spatial mixing}) reduces Day-$120$ ACC from $0.66$ to $0.48$, whereas replacing Non-Local Attention ($\mathcal{A}_{\mathrm{attn}}$) with Dilated Spatial Gating ($\mathcal{A}_{\mathrm{gate}}$) retains strong $120$-day skill ($0.16\,\mathrm{m/s}$ RMSE, $0.61$ ACC) and enables zero-shot cross-resolution transfer across global ocean grids in Section~\ref{sec:exp_rq3}.
\vspace{-6pt}

\section{Conclusion}
\label{sec:conclusion}
\vspace{-5pt}
We presented \method, a scale-spectral decoupled foundation model for long-horizon ocean dynamics. Isolating multi-step evolution on a $4\times$-coarsened latent manifold while routing fine-scale frontal and eddy structures through a single outer bypass and a local scale-spectral composer prevents both high-wavenumber spectral dissipation and autoregressive error accumulation. Extensive evaluations across global $0.25^\circ$, $23$-level 3D ocean simulation and coupled forecasting, $120$-day regional $0.125^\circ$ eddy forecasting, and zero-shot $0.25^\circ\to0.125^\circ$ transfer show that \method outperforms existing ocean models, spatiotemporal backbones, and neural PDE operators in accuracy and spectral fidelity.

\clearpage
\section*{LLM Usage Statement}
Large language models were used solely for language polishing, grammar checking, LaTeX formatting, and assistance with evaluation and visualization scripts during manuscript preparation. No large language models were used to generate research ideas, experimental data, or numerical results. All architectural designs, mathematical derivations, experiments, and scientific claims were developed and independently verified by the authors, who take full responsibility for the integrity and contents of this manuscript.

{\small
\bibliographystyle{ieeenat_fullname}
\bibliography{ocean}
}
\ifdefined\QuantAtlasMainOnly\else
\clearpage
\onecolumn
\appendix
\setcounter{figure}{0}
\renewcommand{\thefigure}{A\arabic{figure}}
\setcounter{table}{0}
\renewcommand{\thetable}{A\arabic{table}}
\setcounter{section}{0}
\renewcommand{\thesection}{\Alph{section}}
\section{Complete Regional Forecast Curves}
\label{app:regional_curves}
Figure~\ref{fig:app_all_curves} compares all twelve baselines, including Persistence, with \method across three western boundary currents. Each curve averages the same 30 initial conditions in 2023 and both velocity components on the unified $0.25^\circ$ scoring grid. No temporal smoothing is applied.
\begin{center}
\includegraphics[width=\linewidth]{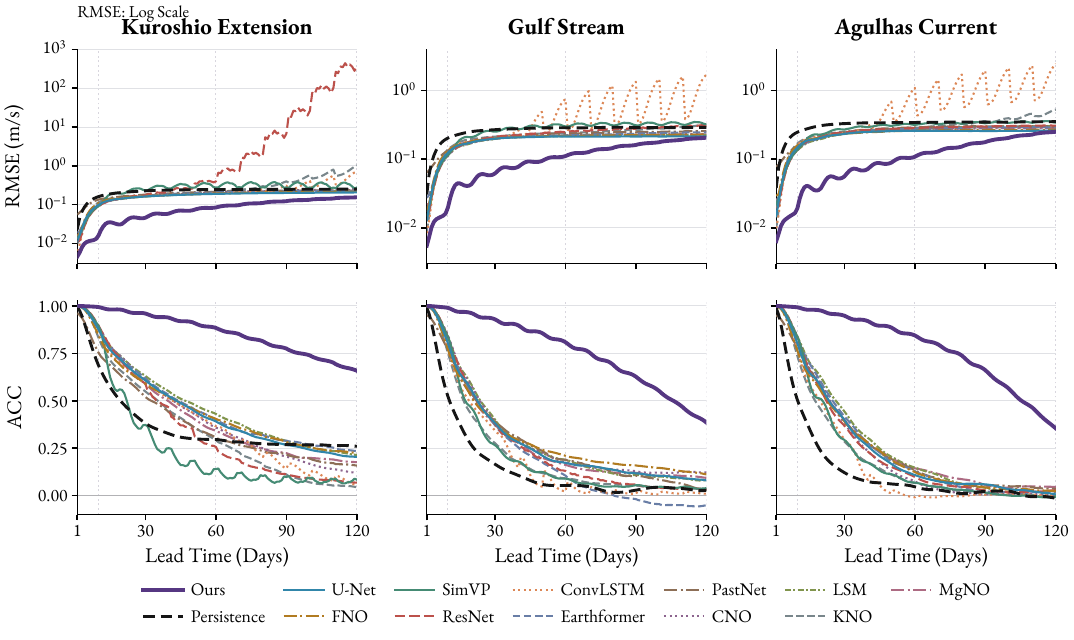}
\captionof{figure}{\textbf{\textit{Complete RMSE and ACC curves over 120 days.}} Columns correspond to the Kuroshio Extension, Gulf Stream, and Agulhas Current. Purple solid curves denote Ours; black dashed curves denote Persistence. RMSE uses logarithmic axes with region-specific ranges to include diverging forecasts; ACC uses a shared linear range.}
\label{fig:app_all_curves}
\end{center}
\noindent\textbf{\textit{Persistence and Metric Definitions.}}
Persistence repeats the last observed state at every lead, $\hat{\mathbf X}_{t+d}=\mathbf X_t$ for $d=1,\ldots,120$. It requires neither training nor external forcing. RMSE and ACC use the same truth-valid ocean mask and cosine-latitude weights as the learned models. For regional ACC, anomalies are measured relative to the 1993 to 2018 per-grid temporal mean. Metrics are first computed separately for each initial condition and velocity component, then averaged over initial conditions and $(u,v)$.

\medskip
\noindent\textbf{\textit{Long-Horizon Interpretation.}}
Persistence provides a useful non-evolving reference. At Day~120 in the Kuroshio, it reaches $0.260$ ACC, exceeding the strongest learned baseline ($0.234$), while \method reaches $0.658$. In the Gulf Stream and Agulhas Current, Persistence falls to $0.029$ and $-0.014$ ACC, respectively. RMSE and ACC provide complementary views: retaining a static spatial pattern need not minimize the velocity error.

\medskip
\noindent\textbf{\textit{Qualitative Comparison Protocol.}}
The following pages enlarge the three regional examples from the main paper. Each uses the first evaluation case, initialized on 1 January 2023, with verification dates of 11 January, 2 March, and 1 May for Days~10, 60, and 120. Reference, Ours, U-Net, and FNO share the same window, grid, ocean mask, speed color scale, and vector scale. Day-120 error maps show $\sqrt{(\hat u-u)^2+(\hat v-v)^2}$ at each ocean point, rather than a spatially averaged RMSE. These are single-case visualizations; the curves above aggregate all 30 cases.
\clearpage
\section{Kuroshio Extension: Flow Structure and Error}
\begin{center}
\includegraphics[width=0.98\linewidth]{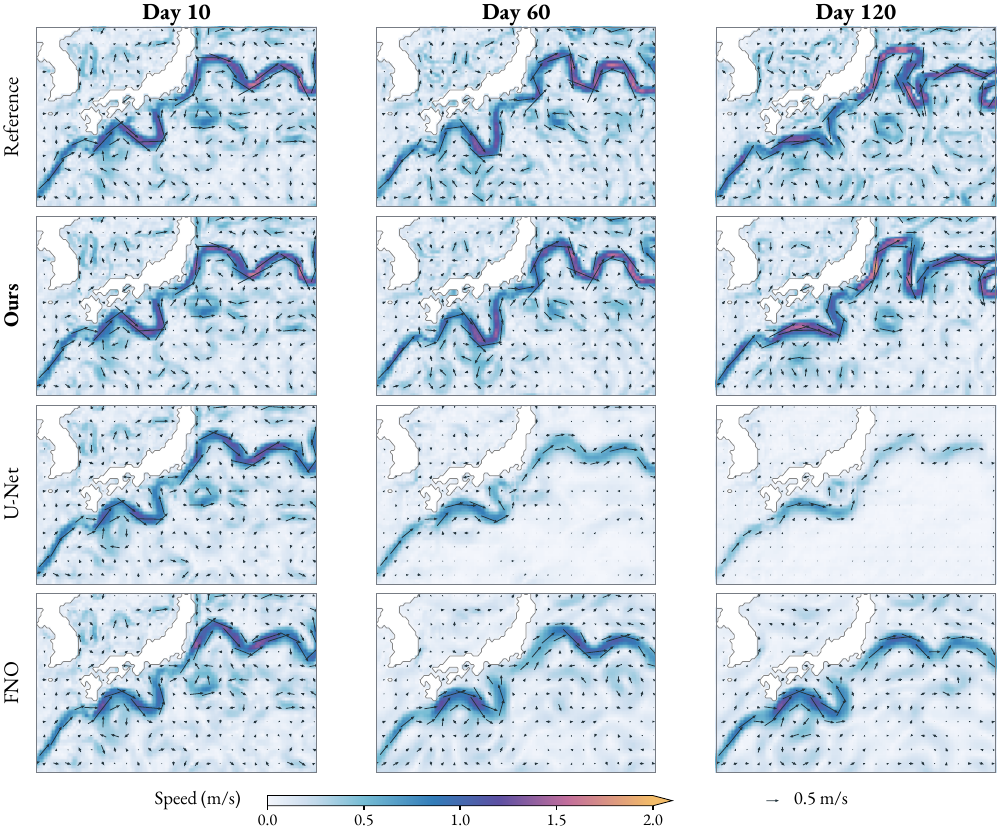}
\captionof{figure}{\textbf{\textit{Kuroshio Extension at Days 10, 60, and 120.}} Rows show Reference, Ours, U-Net, and FNO. Color encodes current speed and arrows show velocity with a common scale. The displayed window spans $125^\circ$E to $154.8^\circ$E, $25^\circ$N to $41^\circ$N; land is white.}
\label{fig:app_kuroshio_speed}
\end{center}
\begin{center}
\includegraphics[width=0.98\linewidth]{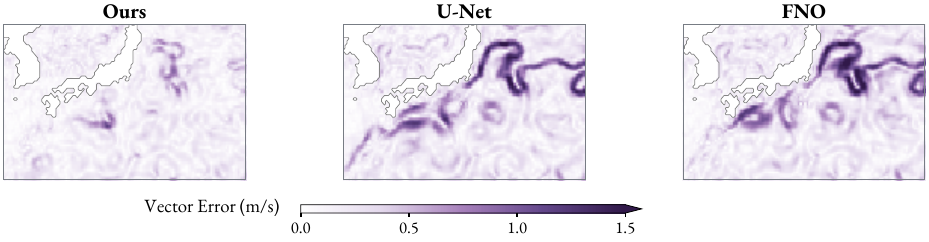}
\captionof{figure}{\textbf{\textit{Day-120 vector error in the Kuroshio Extension.}} Columns show Ours, U-Net, and FNO for the same initial condition and window. All three regions use a shared error color scale; darker colors indicate larger errors.}
\label{fig:app_kuroshio_error}
\end{center}
\clearpage
\section{Gulf Stream: Flow Structure and Error}
\begin{center}
\includegraphics[width=0.98\linewidth]{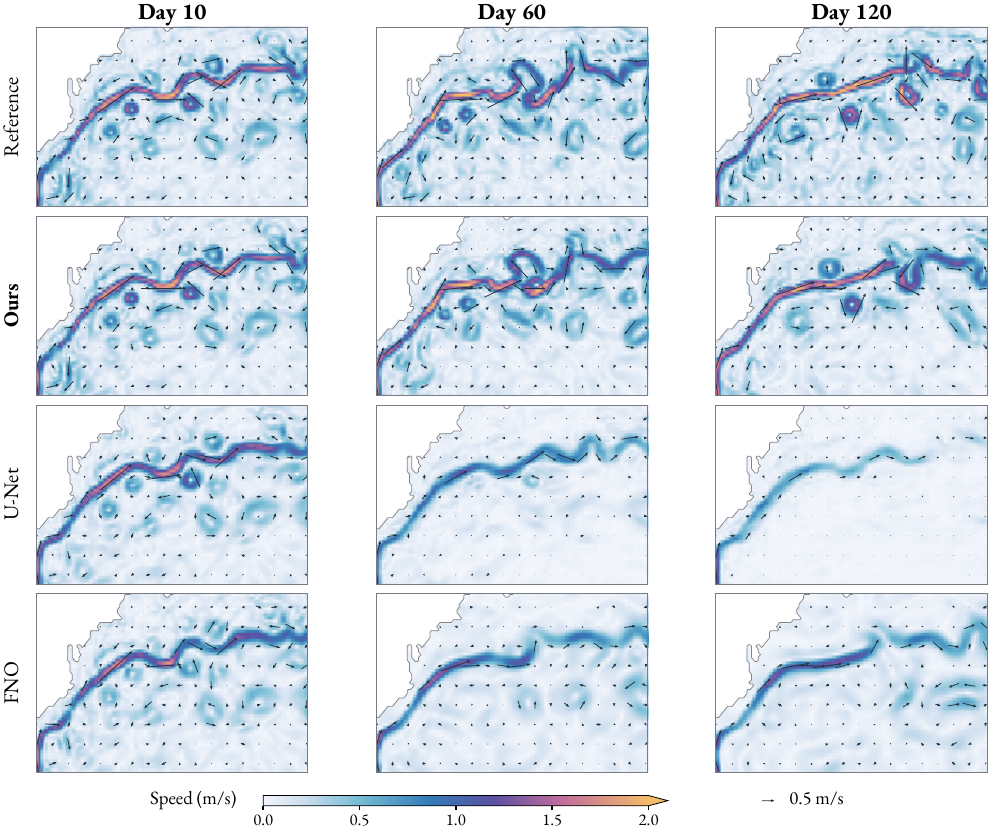}
\captionof{figure}{\textbf{\textit{Gulf Stream at Days 10, 60, and 120.}} Rows show Reference, Ours, U-Net, and FNO. Color encodes current speed and arrows show velocity with a common scale. The displayed window spans $80^\circ$W to $50^\circ$W, $28^\circ$N to $44^\circ$N; land is white.}
\label{fig:app_gulf_speed}
\end{center}
\begin{center}
\includegraphics[width=0.98\linewidth]{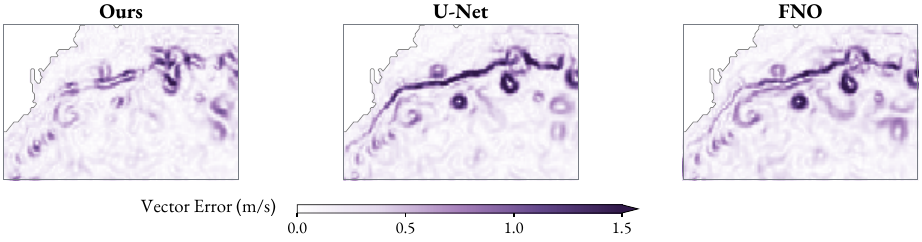}
\captionof{figure}{\textbf{\textit{Day-120 vector error in the Gulf Stream.}} Columns show Ours, U-Net, and FNO for the same initial condition and window. All three regions use a shared error color scale; darker colors indicate larger errors.}
\label{fig:app_gulf_error}
\end{center}
\clearpage
\section{Agulhas Current: Flow Structure and Error}
\begin{center}
\includegraphics[width=0.98\linewidth]{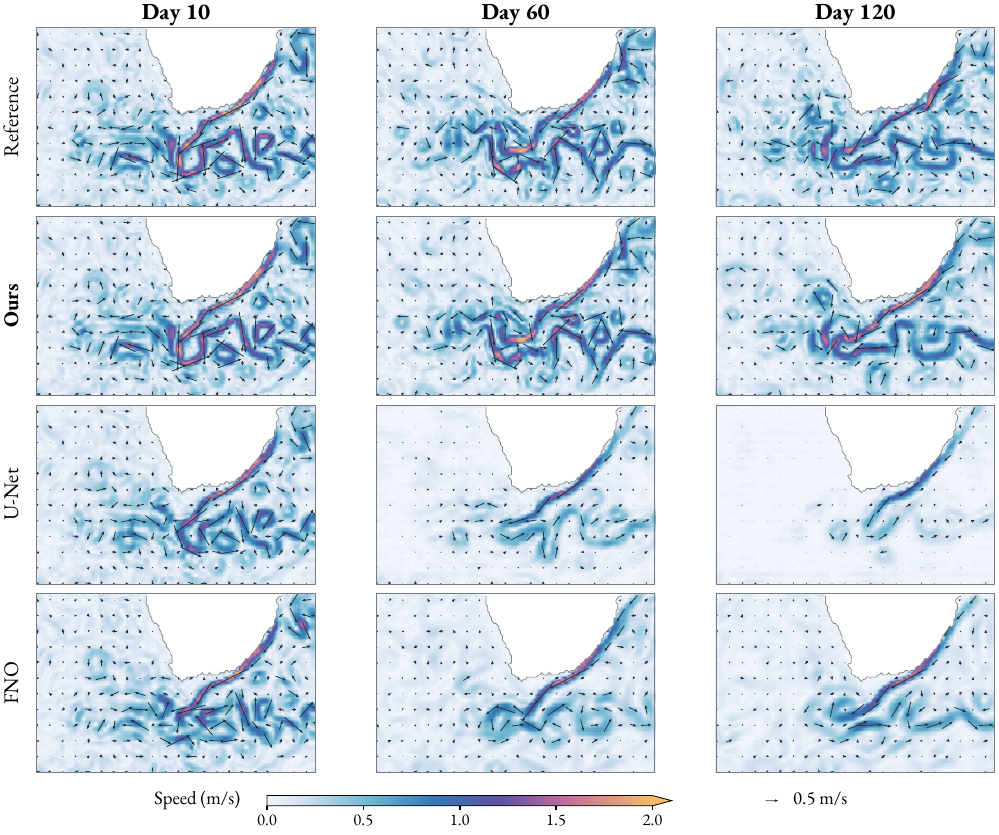}
\captionof{figure}{\textbf{\textit{Agulhas Current at Days 10, 60, and 120.}} Rows show Reference, Ours, U-Net, and FNO. Color encodes current speed and arrows show velocity with a common scale. The displayed window spans $0^\circ$ to $38^\circ$E, $45^\circ$S to $25^\circ$S; land is white.}
\label{fig:app_agulhas_speed}
\end{center}
\begin{center}
\includegraphics[width=0.98\linewidth]{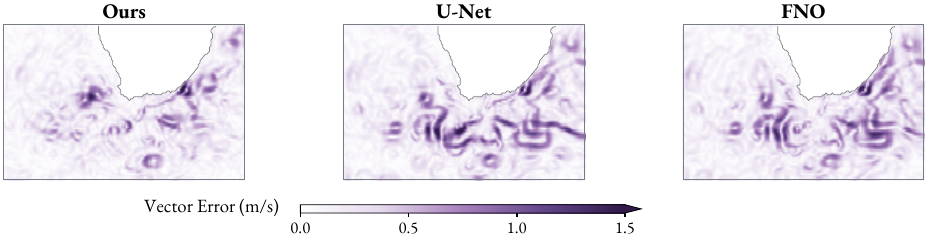}
\captionof{figure}{\textbf{\textit{Day-120 vector error in the Agulhas Current.}} Columns show Ours, U-Net, and FNO for the same initial condition and window. All three regions use a shared error color scale; darker colors indicate larger errors.}
\label{fig:app_agulhas_error}
\end{center}

\clearpage
\section{Global 3D Ocean: Salinity Anomalies}
\begin{center}
\includegraphics[width=0.90\linewidth]{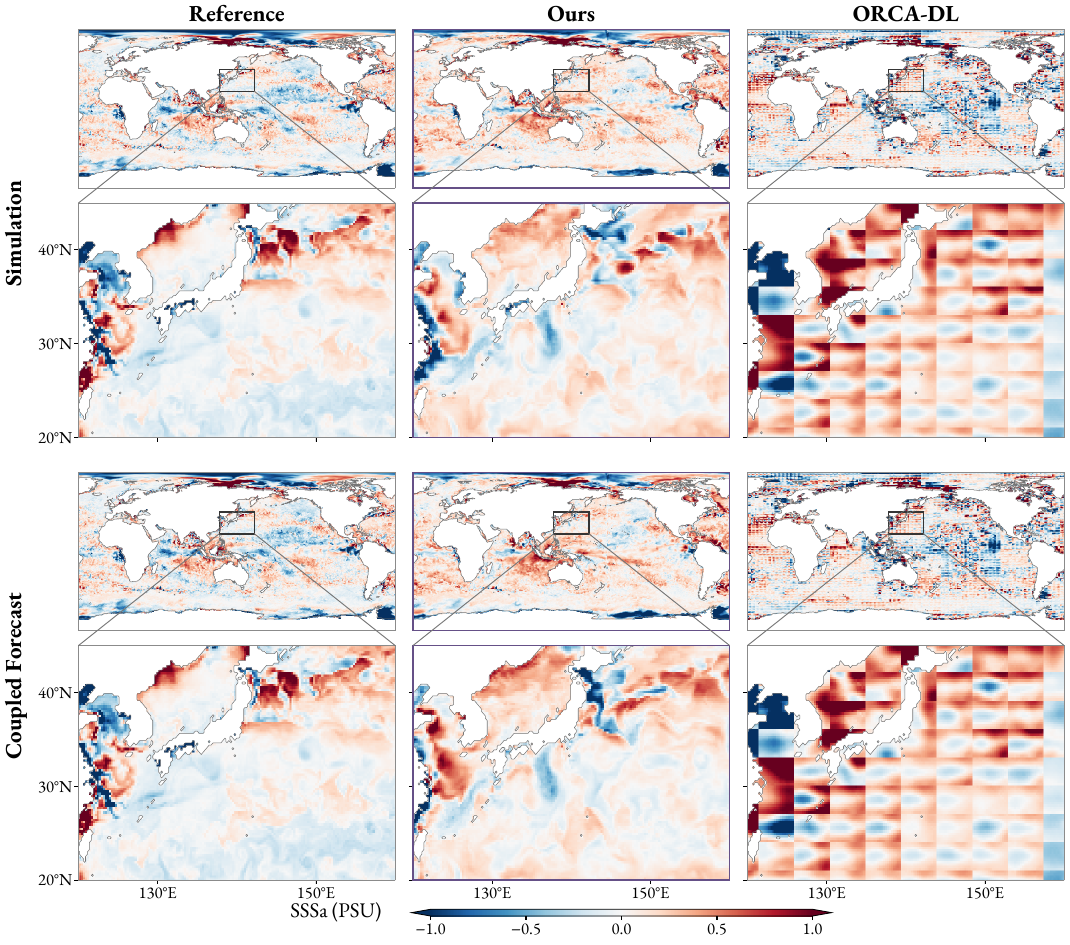}
\captionof{figure}{\textbf{\textit{Sea surface salinity anomalies at Day 60.}} Simulation uses ERA5 forcing; the coupled forecast uses predicted atmospheric forcing. Each column shows the globe and the same $120^\circ$E to $160^\circ$E, $20^\circ$N to $45^\circ$N zoom. All panels share one color scale. Initial date: 1 January 2020; valid date: 1 March 2020. Global display: $0.5^\circ$ sampling; zooms: native $0.25^\circ$.}
\label{fig:app_global3d_sssa}
\end{center}
\begin{center}
\captionof{table}{\textbf{\textit{Single-case global RMSE ($\mathrm{PSU}$).}} Metrics use the full native $0.25^\circ$ ocean grid, cosine-latitude weights, and a common valid mask; they are not 240-case averages. Values use two significant figures; bold denotes the unrounded minimum.}
{\small\setlength{\tabcolsep}{12pt}\renewcommand{\arraystretch}{1.05}
\begin{tabular}{lrrrr}\toprule
Setting & Ours & FourCastNet & AI-GOMS & ORCA-DL\\\midrule
Simulation & \textbf{0.36} & 1.1 & 0.69 & 0.64\\
Coupled forecast & \textbf{0.41} & 1.1 & 0.76 & 0.64\\
\bottomrule\end{tabular}}
\end{center}
\noindent\textbf{\textit{Salinity Structure versus Phase Accuracy.}}
ORCA-DL exhibits repeated, grid-aligned salinity patches in both tasks, whereas Ours produces spatially continuous anomalies. Relative to ORCA-DL, Ours reduces global RMSE by $44\%$ in simulation and $36\%$ in coupled forecasting. The gain is not simply a smaller global offset: Ours retains positive mean biases of $0.106$ and $0.101\,\mathrm{PSU}$. Moreover, its Kuroshio spatial PCC falls from $0.240$ to $-0.022$ between the two tasks. Thus, lower global error and the absence of regular tiling do not establish accurate local anomaly phase; feature displacement remains visible in the zooms.
\clearpage
\section{Global 3D Ocean: Temperature Anomalies}
\begin{center}
\includegraphics[width=0.90\linewidth]{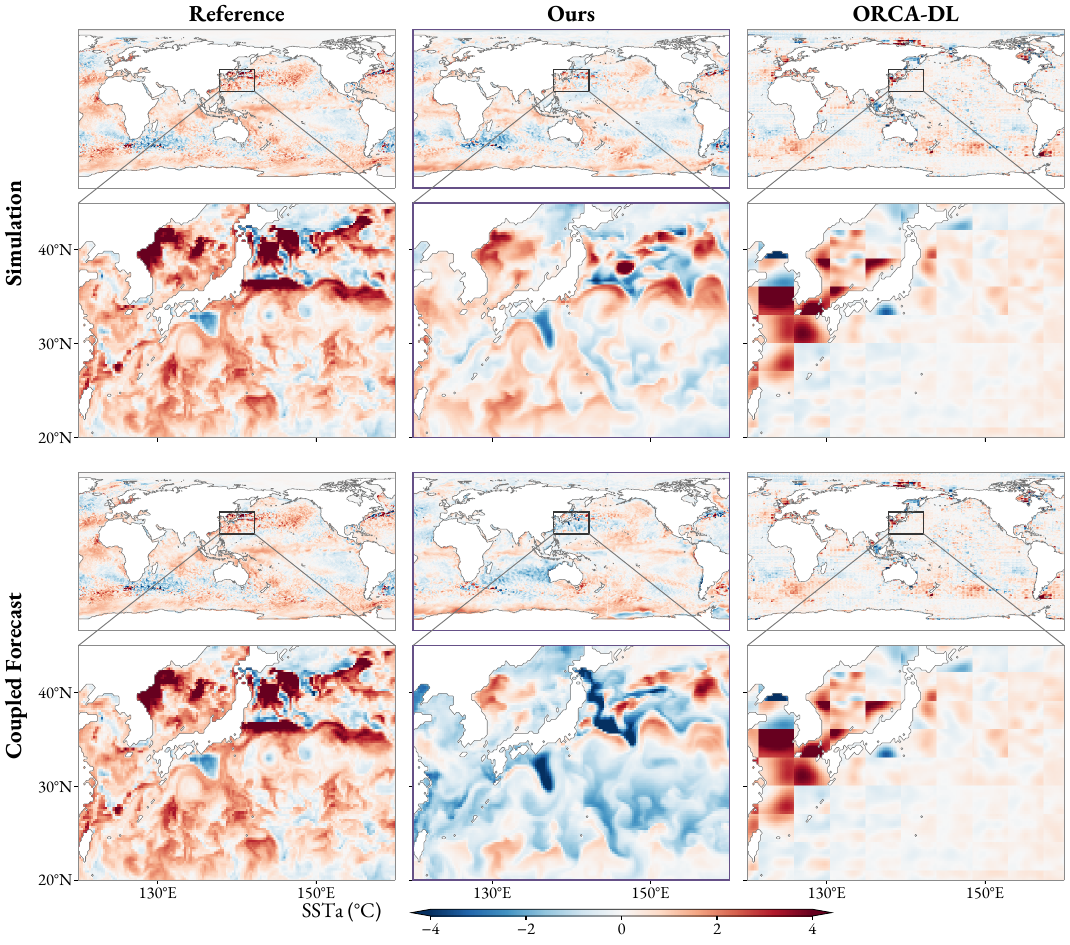}
\captionof{figure}{\textbf{\textit{Sea surface temperature anomalies at Day 60.}} Simulation uses ERA5 forcing; the coupled forecast uses predicted atmospheric forcing. Each column shows the globe and the same $120^\circ$E to $160^\circ$E, $20^\circ$N to $45^\circ$N zoom. All panels share one color scale. Initial date: 1 January 2020; valid date: 1 March 2020. Global display: $0.5^\circ$ sampling; zooms: native $0.25^\circ$.}
\label{fig:app_global3d_ssta}
\end{center}
\begin{center}
\captionof{table}{\textbf{\textit{Single-case global RMSE (${}^\circ\mathrm{C}$).}} Metrics use the full native $0.25^\circ$ ocean grid, cosine-latitude weights, and a common valid mask; they are not 240-case averages. Values use two significant figures; bold denotes the unrounded minimum.}
{\small\setlength{\tabcolsep}{12pt}\renewcommand{\arraystretch}{1.05}
\begin{tabular}{lrrrr}\toprule
Setting & Ours & FourCastNet & AI-GOMS & ORCA-DL\\\midrule
Simulation & \textbf{0.78} & 1.5 & 0.93 & 1.0\\
Coupled forecast & \textbf{1.0} & 1.6 & 1.1 & 1.0\\
\bottomrule\end{tabular}}
\end{center}
\noindent\textbf{\textit{Thermal Bias under Forecast Forcing.}}
Ours captures broad anomaly contrasts in simulation, but the coupled forecast develops a stronger cold anomaly in the Kuroshio window. Its regional RMSE increases from $1.50$ to $2.09\,{}^\circ\mathrm{C}$ and its mean bias changes from $-0.66$ to $-1.41\,{}^\circ\mathrm{C}$. ORCA-DL has a smaller regional forecast RMSE ($1.66\,{}^\circ\mathrm{C}$), despite Ours retaining the lowest global RMSE among the four methods. This contrast separates global accuracy from regional thermal drift and indicates sensitivity to the forcing setting; it does not isolate individual heat-flux or advection errors.
\clearpage
\section{Global 3D Ocean: Meridional Currents}
\begin{center}
\includegraphics[width=0.90\linewidth]{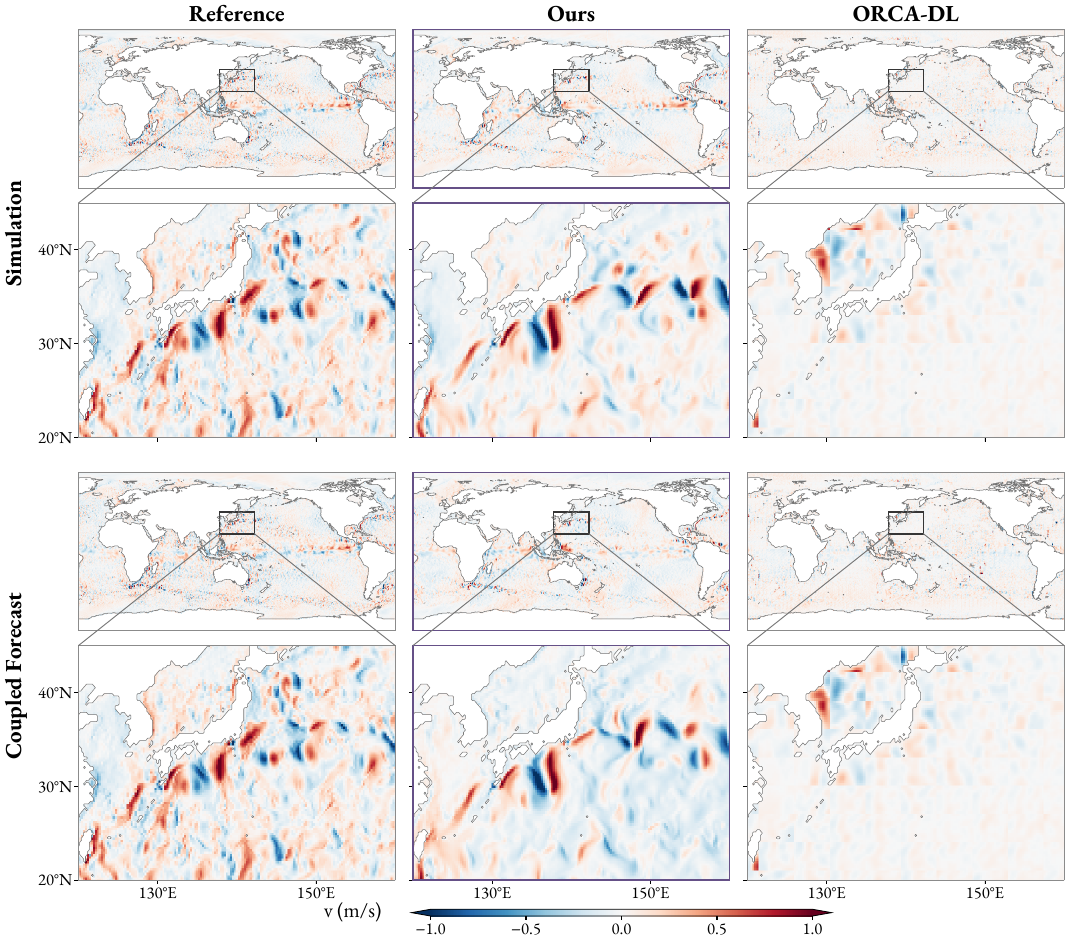}
\captionof{figure}{\textbf{\textit{Surface meridional velocity at Day 60.}} Simulation uses ERA5 forcing; the coupled forecast uses predicted atmospheric forcing. Each column shows the globe and the same $120^\circ$E to $160^\circ$E, $20^\circ$N to $45^\circ$N zoom. All panels share one color scale. Initial date: 1 January 2020; valid date: 1 March 2020. Global display: $0.5^\circ$ sampling; zooms: native $0.25^\circ$.}
\label{fig:app_global3d_va}
\end{center}
\begin{center}
\captionof{table}{\textbf{\textit{Single-case global RMSE ($\mathrm{m/s}$).}} Metrics use the full native $0.25^\circ$ ocean grid, cosine-latitude weights, and a common valid mask; they are not 240-case averages. Values use two significant figures; bold denotes the unrounded minimum.}
{\small\setlength{\tabcolsep}{12pt}\renewcommand{\arraystretch}{1.05}
\begin{tabular}{lrrrr}\toprule
Setting & Ours & FourCastNet & AI-GOMS & ORCA-DL\\\midrule
Simulation & \textbf{0.15} & 0.46 & 0.16 & 0.18\\
Coupled forecast & \textbf{0.17} & 0.44 & 0.18 & 0.18\\
\bottomrule\end{tabular}}
\end{center}
\noindent\textbf{\textit{Current Structure and Amplitude Error.}}
The Kuroshio zooms show alternating northward and southward velocity lobes in the reference and Ours; these lobes are strongly attenuated in ORCA-DL. For the coupled forecast, the regional spatial PCC is $0.422$ for Ours versus $0.082$ for ORCA-DL, although their regional RMSE values are nearly identical ($0.223\,\mathrm{m/s}$). This demonstrates why amplitude error alone does not characterize circulation structure. Ours still shows displaced or intensified lobes, so retaining visible variability should not be interpreted as exact eddy tracking or evidence of subsurface dynamical fidelity.

\smallskip\noindent These diagnostics concern surface outputs of the 23-level model. They complement the multi-initial-condition scores and do not assess vertical profiles or three-dimensional transport budgets.

\clearpage
\section{Global 3D Ocean: Sea Level Anomalies}
\begin{center}
\includegraphics[width=0.90\linewidth]{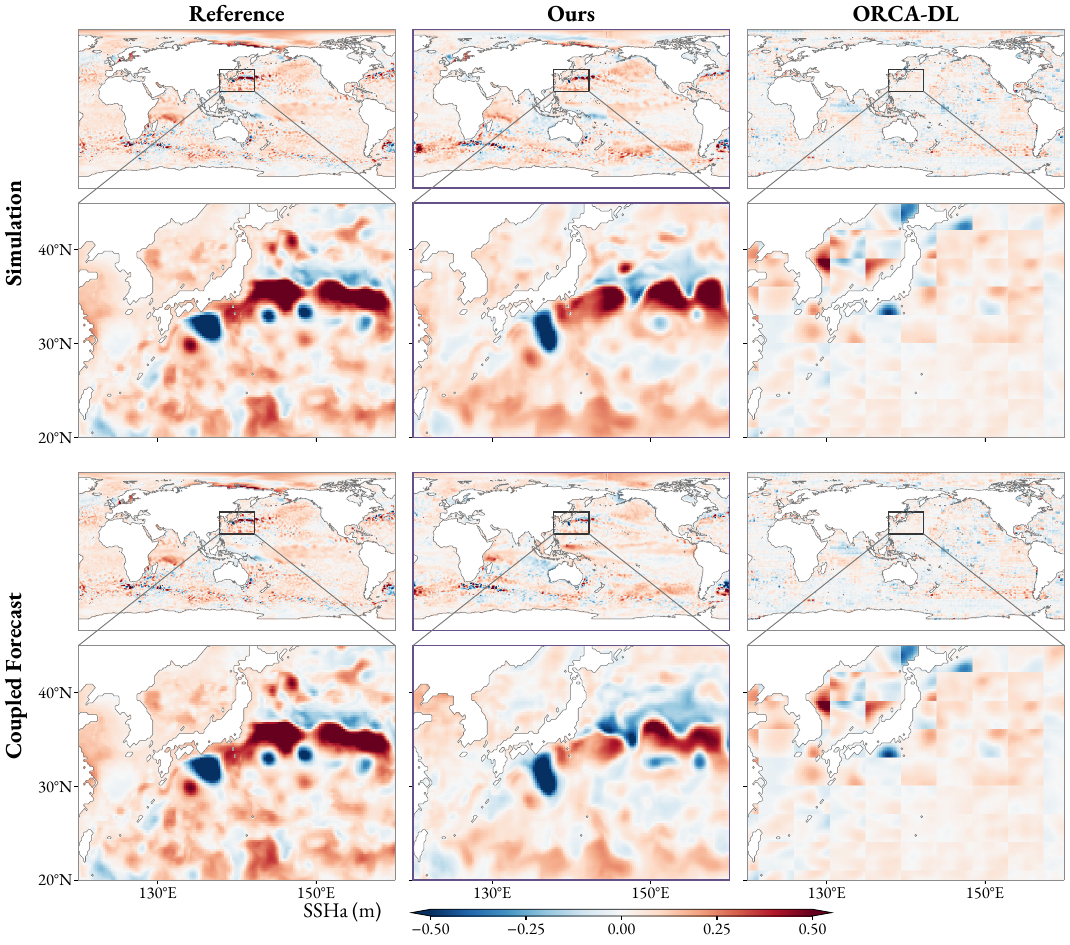}
\captionof{figure}{\textbf{\textit{Sea surface height anomalies at Day 60.}} Rows compare simulation and coupled forecasting; columns use Reference, Ours, and the same ORCA-DL comparator. Global maps use $0.5^\circ$ sampling and black rectangles locate the native $0.25^\circ$ Kuroshio zooms. All panels share one color scale; extensions indicate saturation. Initial date: 1 January 2020; valid date: 1 March 2020.}
\label{fig:app_global3d_ssha}
\end{center}
\begin{center}
\captionof{table}{\textbf{\textit{Single-case global RMSE ($\mathrm{m}$).}} Metrics use the native $0.25^\circ$ grid, cosine-latitude weights, and a common valid ocean mask. Values use two significant figures; bold denotes the unrounded minimum.}
{\small\setlength{\tabcolsep}{12pt}\renewcommand{\arraystretch}{1.05}
\begin{tabular}{lrrrr}\toprule
Setting & Ours & FourCastNet & AI-GOMS & ORCA-DL\\\midrule
Simulation & \textbf{0.093} & 0.46 & 0.098 & 0.12\\
Coupled forecast & 0.11 & 0.44 & \textbf{0.11} & 0.12\\
\bottomrule\end{tabular}}
\end{center}
\noindent\textbf{\textit{Sea Level Pattern and Forcing Sensitivity.}}
Sea level anomalies provide a complementary view of the large-scale pressure field and mesoscale structure. Ours has a global simulation RMSE of $0.093\,\mathrm{m}$, which increases to $0.111\,\mathrm{m}$ under forecast forcing. In the Kuroshio window, spatial PCC decreases from $0.614$ to $0.497$, while the mean bias changes from $-0.007$ to $-0.049\,\mathrm{m}$. The paired panels therefore distinguish anomaly-pattern retention from a forcing-dependent offset. AI-GOMS has the lowest global forecast RMSE in this case ($0.108\,\mathrm{m}$); the simulation advantage does not extend uniformly across settings.

\clearpage
\section{Global 3D Ocean: Zonal Currents}
\begin{center}
\includegraphics[width=0.90\linewidth]{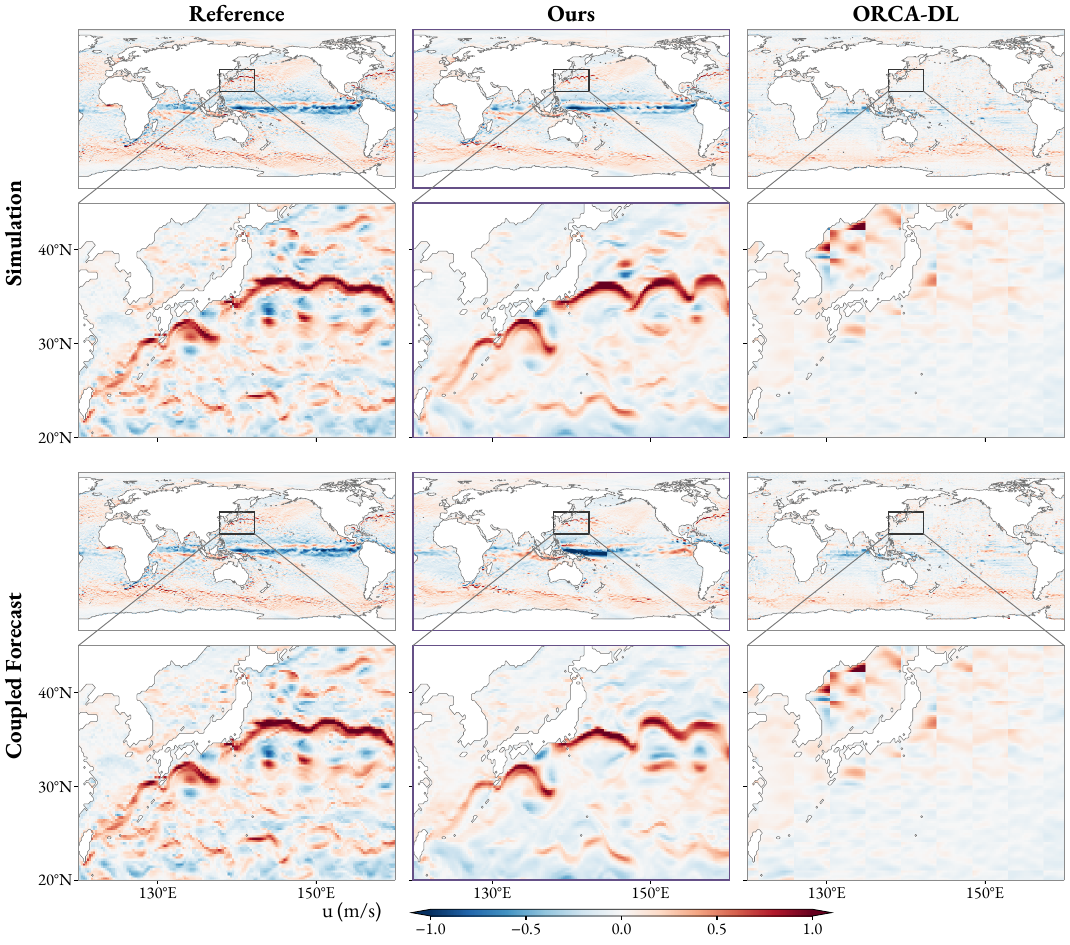}
\captionof{figure}{\textbf{\textit{Surface zonal velocity at Day 60.}} Rows compare simulation and coupled forecasting; columns use Reference, Ours, and the same ORCA-DL comparator. Global maps use $0.5^\circ$ sampling and black rectangles locate the native $0.25^\circ$ Kuroshio zooms. All panels share one color scale; extensions indicate saturation. Initial date: 1 January 2020; valid date: 1 March 2020.}
\label{fig:app_global3d_ua}
\end{center}
\begin{center}
\captionof{table}{\textbf{\textit{Single-case global RMSE ($\mathrm{m/s}$).}} Metrics use the native $0.25^\circ$ grid, cosine-latitude weights, and a common valid ocean mask. Values use two significant figures; bold denotes the unrounded minimum.}
{\small\setlength{\tabcolsep}{12pt}\renewcommand{\arraystretch}{1.05}
\begin{tabular}{lrrrr}\toprule
Setting & Ours & FourCastNet & AI-GOMS & ORCA-DL\\\midrule
Simulation & \textbf{0.16} & 0.46 & 0.17 & 0.23\\
Coupled forecast & 0.22 & 0.46 & \textbf{0.21} & 0.23\\
\bottomrule\end{tabular}}
\end{center}
\noindent\textbf{\textit{Zonal Flow Structure.}}
Positive and negative velocities distinguish eastward and westward flow, allowing the current branches to be compared without collapsing direction into speed. Ours has a global RMSE of $0.157\,\mathrm{m/s}$ in simulation and $0.216\,\mathrm{m/s}$ in the coupled forecast. Its Kuroshio spatial PCC decreases from $0.478$ to $0.417$, indicating only partial agreement in the position and strength of the regional flow. AI-GOMS achieves a lower global forecast RMSE ($0.207\,\mathrm{m/s}$). These diagnostics complement the meridional component: a convincing current reconstruction requires both components to be accurate.

\clearpage
\section{Global 3D Ocean: Salinity at 109 m}
\begin{center}
\includegraphics[width=0.90\linewidth]{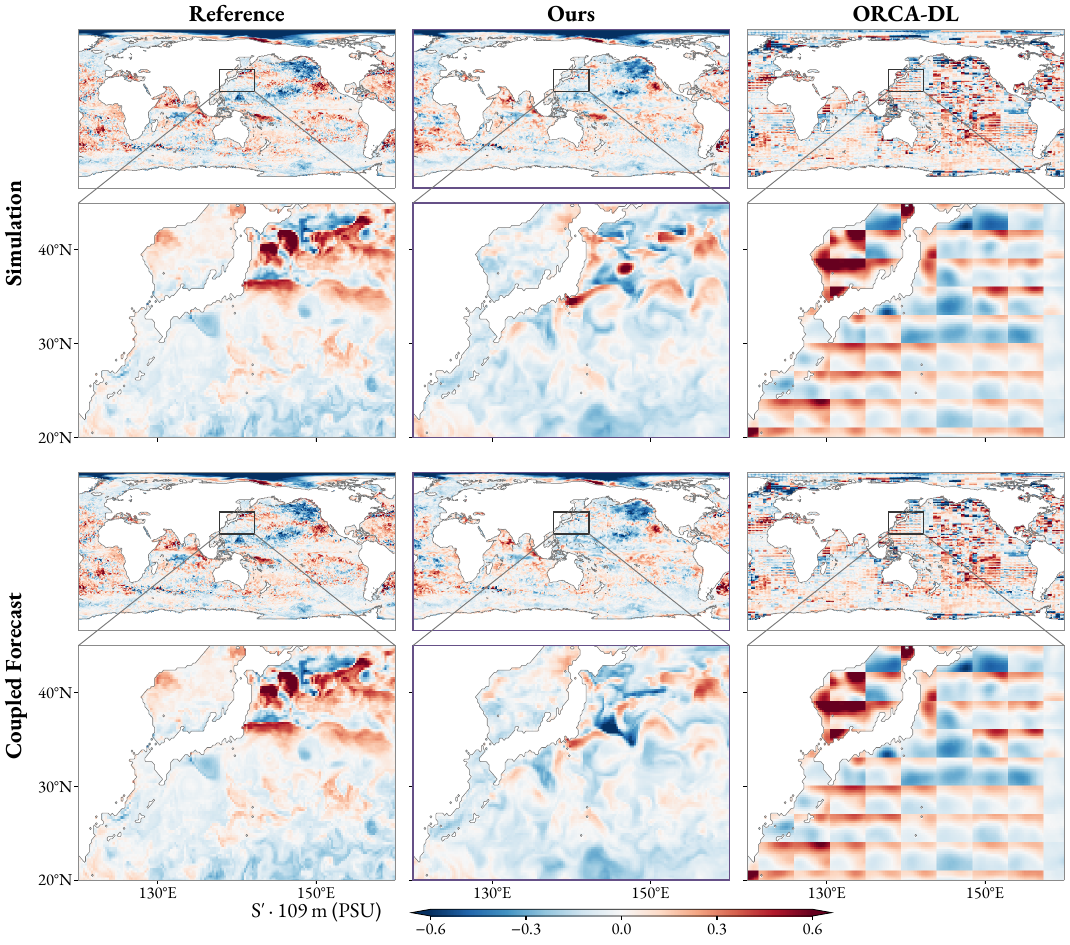}
\captionof{figure}{\textbf{\textit{Salinity anomalies at a nominal depth of 109 m at Day 60.}} Rows compare simulation and coupled forecasting; columns use Reference, Ours, and the same ORCA-DL comparator. Global maps use $0.5^\circ$ sampling and black rectangles locate the native $0.25^\circ$ Kuroshio zooms. All panels share one color scale; extensions indicate saturation. Initial date: 1 January 2020; valid date: 1 March 2020.}
\label{fig:app_global3d_s109}
\end{center}
\begin{center}
\captionof{table}{\textbf{\textit{Single-case global RMSE ($\mathrm{PSU}$).}} Metrics use the native $0.25^\circ$ grid, cosine-latitude weights, and a common valid ocean mask. Values use two significant figures; bold denotes the unrounded minimum.}
{\small\setlength{\tabcolsep}{12pt}\renewcommand{\arraystretch}{1.05}
\begin{tabular}{lrrrr}\toprule
Setting & Ours & FourCastNet & AI-GOMS & ORCA-DL\\\midrule
Simulation & \textbf{0.16} & 4.1 & 0.32 & 0.29\\
Coupled forecast & \textbf{0.17} & 3.8 & 0.33 & 0.29\\
\bottomrule\end{tabular}}
\end{center}
\noindent\textbf{\textit{Subsurface Salinity Pattern.}}
At 109 m, Ours maintains similar global RMSE under observed and predicted forcing ($0.162$ and $0.165\,\mathrm{PSU}$), both below the three baselines in this case. Nevertheless, its Kuroshio spatial PCC is only $0.240$ in simulation and $0.185$ in forecasting. Thus, a small change in global error does not imply that the localized anomaly pattern is well aligned. The subsurface comparison extends the surface diagnostics to a depth where temperature and salinity jointly influence density, but these anomaly maps alone do not quantify stratification or mixing.

\clearpage
\section{Global 3D Ocean: Temperature at 109 m}
\begin{center}
\includegraphics[width=0.90\linewidth]{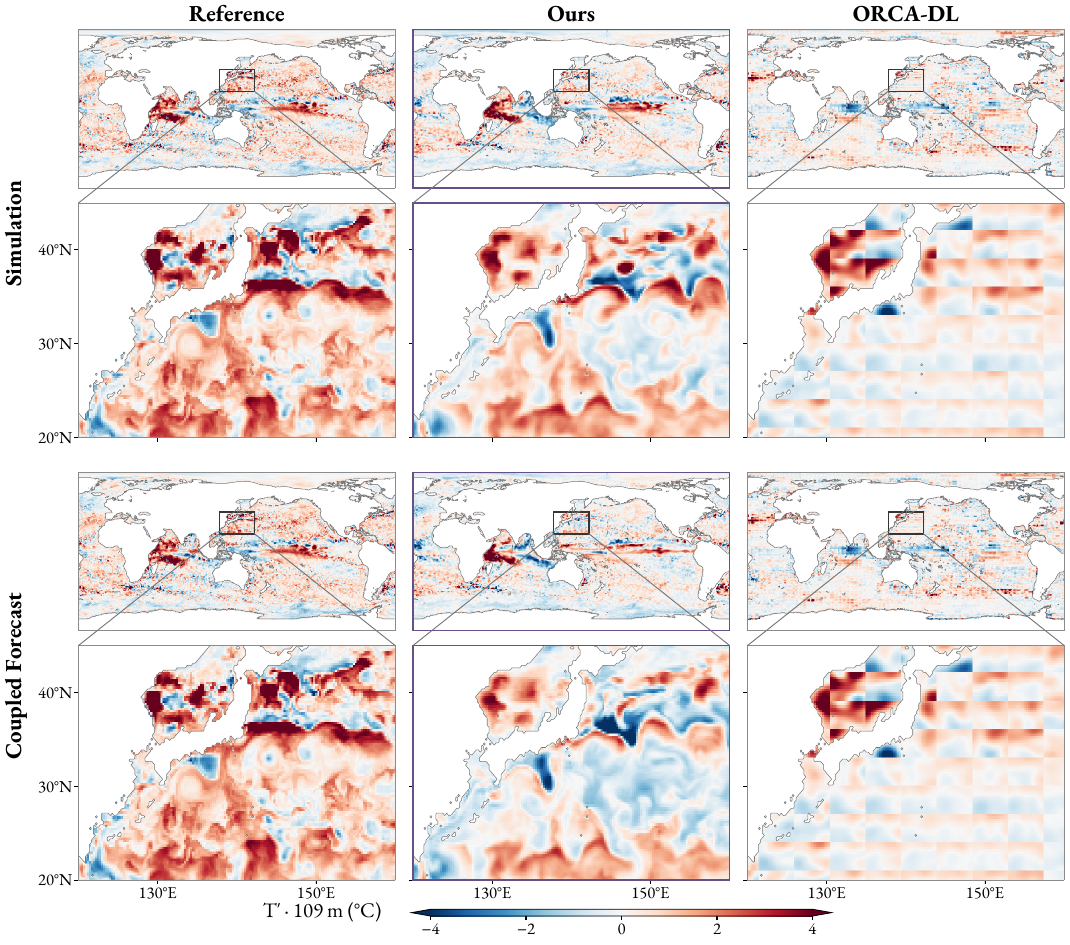}
\captionof{figure}{\textbf{\textit{Temperature anomalies at a nominal depth of 109 m at Day 60.}} Rows compare simulation and coupled forecasting; columns use Reference, Ours, and the same ORCA-DL comparator. Global maps use $0.5^\circ$ sampling and black rectangles locate the native $0.25^\circ$ Kuroshio zooms. All panels share one color scale; extensions indicate saturation. Initial date: 1 January 2020; valid date: 1 March 2020.}
\label{fig:app_global3d_t109}
\end{center}
\begin{center}
\captionof{table}{\textbf{\textit{Single-case global RMSE (${}^\circ\mathrm{C}$).}} Metrics use the native $0.25^\circ$ grid, cosine-latitude weights, and a common valid ocean mask. Values use two significant figures; bold denotes the unrounded minimum. One nonfinite reference cell is excluded consistently for all models.}
{\small\setlength{\tabcolsep}{12pt}\renewcommand{\arraystretch}{1.05}
\begin{tabular}{lrrrr}\toprule
Setting & Ours & FourCastNet & AI-GOMS & ORCA-DL\\\midrule
Simulation & \textbf{1.1} & 5.2 & 1.4 & 1.4\\
Coupled forecast & \textbf{1.3} & 5.2 & 1.4 & 1.4\\
\bottomrule\end{tabular}}
\end{center}
\noindent\textbf{\textit{Subsurface Thermal Drift.}}
Ours has global RMSE values of $1.14$ and $1.28\,{}^\circ\mathrm{C}$ in simulation and forecasting. The forcing contrast is stronger in the Kuroshio window: RMSE increases from $1.58$ to $1.83\,{}^\circ\mathrm{C}$ and the cold bias deepens from $-0.50$ to $-0.89\,{}^\circ\mathrm{C}$. The regional spatial PCC also decreases from $0.362$ to $0.272$. A cold bias therefore occurs below the surface as well as at the surface in this case. Identifying whether it arises from advection, vertical exchange, or forcing errors requires additional budget diagnostics.

\clearpage
\section{Global 3D Ocean: Salinity at 453 m}
\begin{center}
\includegraphics[width=0.90\linewidth]{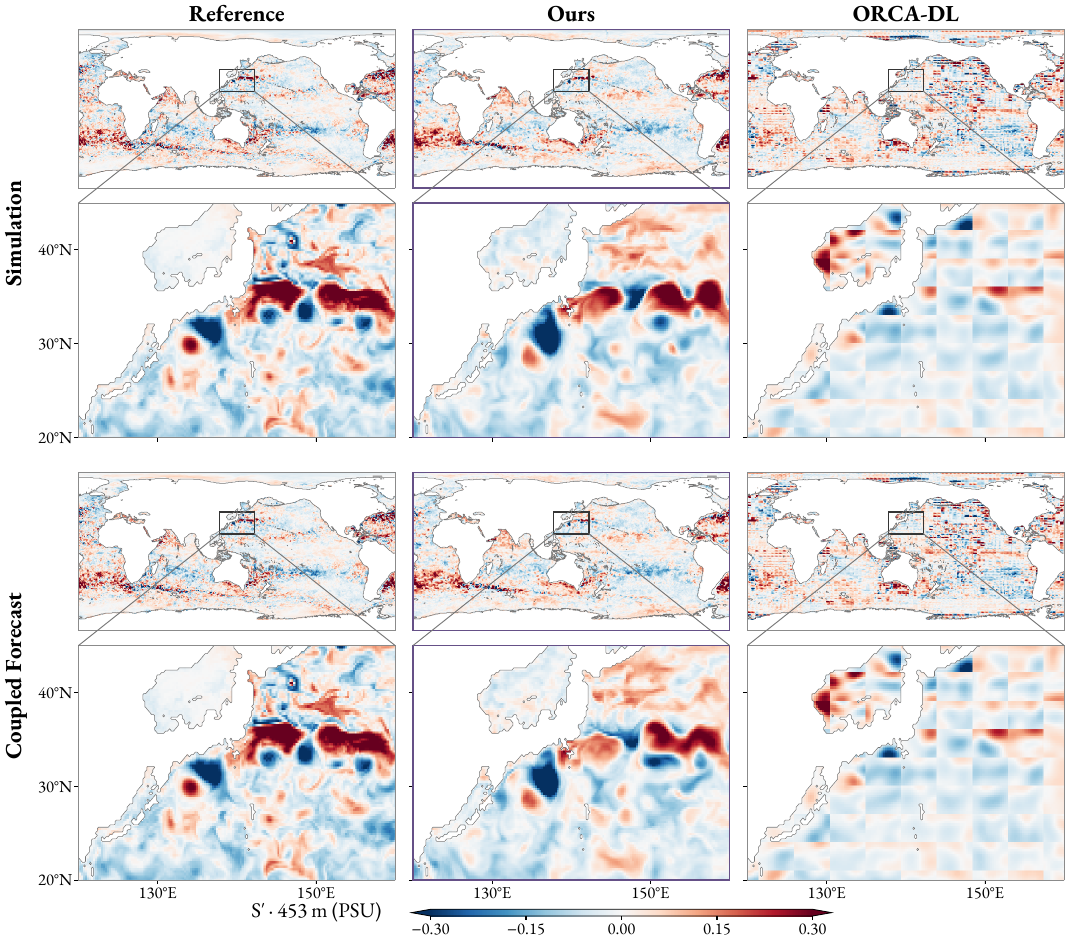}
\captionof{figure}{\textbf{\textit{Salinity anomalies at a nominal depth of 453 m at Day 60.}} Rows compare simulation and coupled forecasting; columns use Reference, Ours, and the same ORCA-DL comparator. Global maps use $0.5^\circ$ sampling and black rectangles locate the native $0.25^\circ$ Kuroshio zooms. All panels share one color scale; extensions indicate saturation. Initial date: 1 January 2020; valid date: 1 March 2020.}
\label{fig:app_global3d_s453}
\end{center}
\begin{center}
\captionof{table}{\textbf{\textit{Single-case global RMSE ($\mathrm{PSU}$).}} Metrics use the native $0.25^\circ$ grid, cosine-latitude weights, and a common valid ocean mask. Values use two significant figures; bold denotes the unrounded minimum.}
{\small\setlength{\tabcolsep}{12pt}\renewcommand{\arraystretch}{1.05}
\begin{tabular}{lrrrr}\toprule
Setting & Ours & FourCastNet & AI-GOMS & ORCA-DL\\\midrule
Simulation & \textbf{0.095} & 0.43 & 0.13 & 0.13\\
Coupled forecast & \textbf{0.096} & 0.41 & 0.13 & 0.13\\
\bottomrule\end{tabular}}
\end{center}
\noindent\textbf{\textit{Depth Dependence of Salinity Error.}}
At 453 m, the global salinity RMSE of Ours changes little between simulation and forecasting ($0.0951$ versus $0.0959\,\mathrm{PSU}$). In the Kuroshio window, its mean bias remains close to zero, but spatial PCC decreases from $0.531$ to $0.440$. This separates a small domain-mean offset from errors in the placement of positive and negative anomalies. The smaller absolute RMSE than at 109 m should not be interpreted as greater relative skill without accounting for the depth-dependent signal variance. The panels retain the original horizontal grid and do not interpolate a vertical section.

\clearpage
\section{Global 3D Ocean: Temperature at 453 m}
\begin{center}
\includegraphics[width=0.90\linewidth]{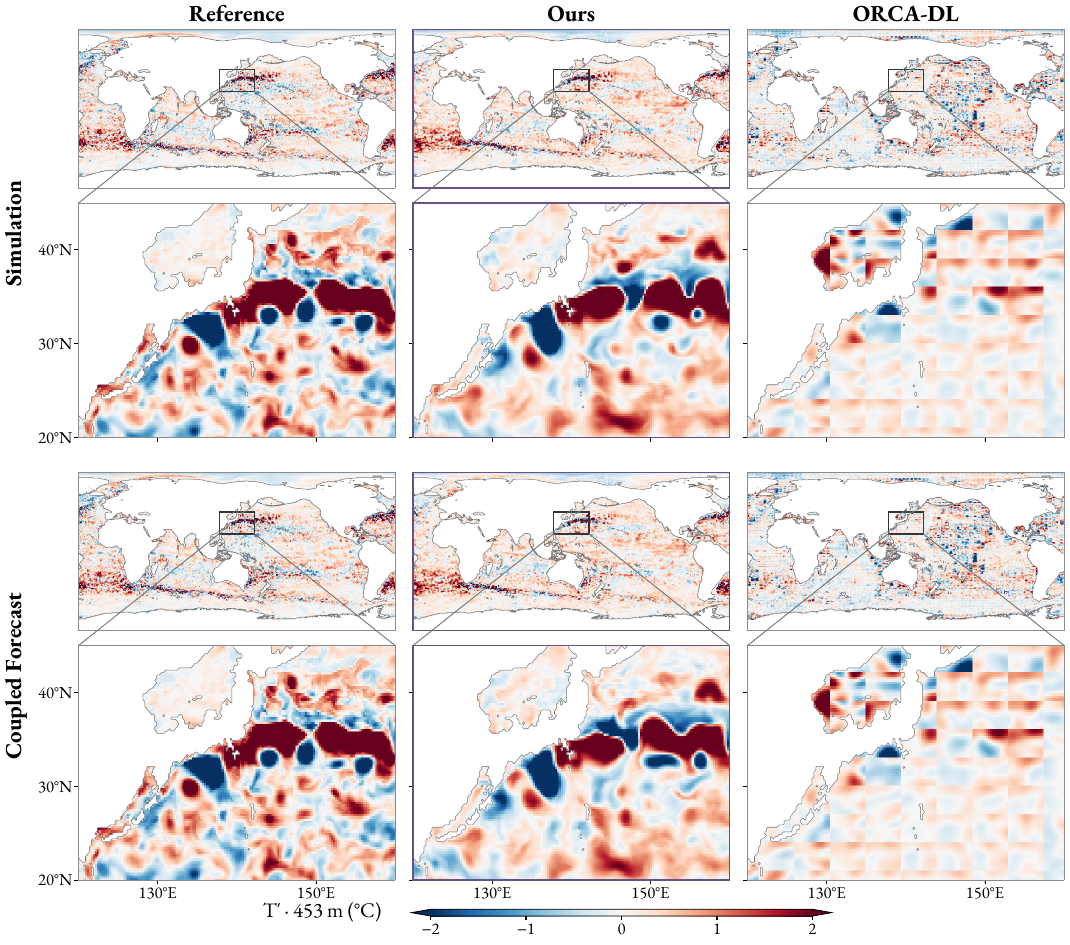}
\captionof{figure}{\textbf{\textit{Temperature anomalies at a nominal depth of 453 m at Day 60.}} Rows compare simulation and coupled forecasting; columns use Reference, Ours, and the same ORCA-DL comparator. Global maps use $0.5^\circ$ sampling and black rectangles locate the native $0.25^\circ$ Kuroshio zooms. All panels share one color scale; extensions indicate saturation. Initial date: 1 January 2020; valid date: 1 March 2020.}
\label{fig:app_global3d_t453}
\end{center}
\begin{center}
\captionof{table}{\textbf{\textit{Single-case global RMSE (${}^\circ\mathrm{C}$).}} Metrics use the native $0.25^\circ$ grid, cosine-latitude weights, and a common valid ocean mask. Values use two significant figures; bold denotes the unrounded minimum.}
{\small\setlength{\tabcolsep}{12pt}\renewcommand{\arraystretch}{1.05}
\begin{tabular}{lrrrr}\toprule
Setting & Ours & FourCastNet & AI-GOMS & ORCA-DL\\\midrule
Simulation & \textbf{0.69} & 2.3 & 0.79 & 0.86\\
Coupled forecast & \textbf{0.71} & 2.3 & 0.8 & 0.86\\
\bottomrule\end{tabular}}
\end{center}
\noindent\textbf{\textit{Thermal Structure at Depth.}}
At 453 m, the global temperature RMSE of Ours increases from $0.690$ to $0.707\,{}^\circ\mathrm{C}$ under forecast forcing, a smaller absolute increase than at 109 m. Its Kuroshio spatial PCC remains $0.612$ in simulation and $0.520$ in forecasting, while regional RMSE increases from $1.38$ to $1.50\,{}^\circ\mathrm{C}$. The retained anomaly pattern and remaining local errors are both visible in the zooms. Together with the 109 m panels, this comparison characterizes depth-dependent forecast behavior at two saved levels; it does not establish accurate vertical heat transport or a closed heat budget.

\fi
\end{document}